\documentclass[a4paper,fleqn]{cas-dc}

\usepackage{placeins}
\usepackage{float}
\usepackage{amsmath}
\usepackage{enumitem}
\usepackage{threeparttable}
\usepackage{subcaption}
\usepackage{pifont}
\usepackage{xcolor}
\usepackage{annotate-equations} 
\usepackage{caption}

\newcommand{\blue}[1]{{\leavevmode\color{blue}#1}}

\usepackage[numbers,sort&compress]{natbib}

\begin{document}
\let\printorcid\relax
\let\WriteBookmarks\relax
\renewcommand{\topfraction}{0.95}
\renewcommand{\bottomfraction}{0.95}
\renewcommand{\textfraction}{0.05}
\renewcommand{\floatpagefraction}{0.9}

\shortauthors{J.H. Huang et~al.}
\shorttitle{}

\title [mode = title]{BeltCrack: the First Sequential-image Conveyor Belt Crack Detection Dataset and Its Baseline with Triple-domain Feature Learning }
\tnotemark[1]

\tnotetext[1]{This work is supported by the
National Natural Science Foundation of China (NSFC) under Grants 62476049 and 62276048.}

\author[1]{Jianghong Huang}
\ead{jianghong@std.uestc.edu.cn}

\affiliation[1]{organization={School of Computer Science and Engineering, University of Electronic Science and Technology of China},
                city={Chengdu},
                postcode={611731}, 
                state={Sichuan},
                country={China}}

\author[1]{Luping Ji}
\cormark[1]
\ead{jiluping@uestc.edu.cn}

\author[2]{Xin Ma}
\ead{hongdunalarm@163.com}

\affiliation[2]{organization={Suzhou Red Shield Intelligent Manufacturing Co., Ltd.},
                city={Suzhou},
                postcode={215000}, 
                country={China}}

\author[1]{Mao Ye}
\ead{cvlab.uestc@gmail.com}

\cortext[1]{Corresponding author}

\begin{abstract}
Conveyor belt, as one of the most widely-used equipment in modern industry, is critical for ensuring both production safety and efficiency. In industrial scenarios, due to long-term running, conveyor belts could often be torn, resulting in cracks. Therefore, how to timely and accurately detect existing belt cracks has significant application potential.
To advance intelligent belt crack detection, sufficient  samples are often crucial. However, existing datasets are mainly for pavement scenarios or synthetic crack data. There has hardly been a real-world and large-scale crack dataset of industrial conveyor belts yet, available for training and testing detection models. Aiming to propel this research field by machine learning, we construct the first pair of sequential-image belt crack detection datasets (BeltCrack14ks with 14,087 images from 29 sequences, and BeltCrack9kd with 9,645 images from 42 sequences) from real-world factory scenes. Furthermore, to validate their usability and effectiveness, we further propose a learning-based baseline method specially designed for both datasets. In our baseline, it incorporates frequency-domain feature learning into temporal-spatial feature fusion for the first time. Moreover, a refined residual mechanism is designed to enable hierarchical feature aggregation. Experiments demonstrate the availability and effectiveness of our two new belt crack datasets. At the same time, they also show that our baseline is significantly superior to other general object detection methods.
Our datasets and source codes are available at \blue{https://github.com/UESTC-nnLab/BeltCrack}.

\end{abstract}


\maketitle
\section{Introduction}
\label{sec1}
Industrial conveyor belts are fundamental to continuous raw material transport in modern production systems \cite{CHEN2023109321}. Their stable operation is critical to production efficiency and worker safety. While stable conveying ensures smooth workflow and productivity, equipment failure can cause downtime, material accumulation, or severe industrial accidents\cite{Peng2024}. Among various health risks, belt cracks caused by foreign intrusion or aging remain a major safety hazard requiring accurate detection \cite{BERMEOAYERBE2023106625}.

To mitigate these risks, early and reliable crack detection has become a key concern \cite{Zhang2023}. Prevailing approach focuses on early detection to prevent crack propagation and catastrophic belt breakage,
thereby enhancing operational safety and reducing economic losses from unplanned downtime and repairs \cite{Zhang2025}. Consequently, achieving reliable detection is increasingly urgent. In this context, vision-based machine learning methods have emerged as promising solutions due to their fast, and non-invasive characteristics, enabling efficient inspection with minimal operational disruption \citep{FANG2020107474}.

However, learning-based belt crack detection methods remains under-explored, primarily due to the scarcity of real-world datasets. Many existing datasets are either synthetic (e.g., MBTID \citep{Wang2022} and CBCD \citep{Dwivedi2023}) or domain-irrelevant (e.g., CrackSeg5k \citep{Chen2024CVPR}). This scarcity  stems from two primary constraints: industrial confidentiality requirements and harsh operational conditions (e.g., high heat or radiation).

To address the lack of real-world datasets for industrial conveyor belt crack detection and to further advance learning-based research in this field, we construct a pair of sequential-image datasets, BeltCrack14ks and BeltCrack9kd, specifically designed for belt crack detection.
Unlike synthetic datasets, both are collected using video cameras on operating conveyor lines in a steel enterprise under real production conditions. Each image is manually annotated through a time-consuming labeling process to ensure accuracy and consistency, with additional verification such as cross-checking and spot validation to maintain annotation quality.
To the best of our knowledge, these are the first real-world industrial datasets dedicated to belt crack detection.

\begin{figure}
    \centering
	\includegraphics[width=1.0\linewidth]{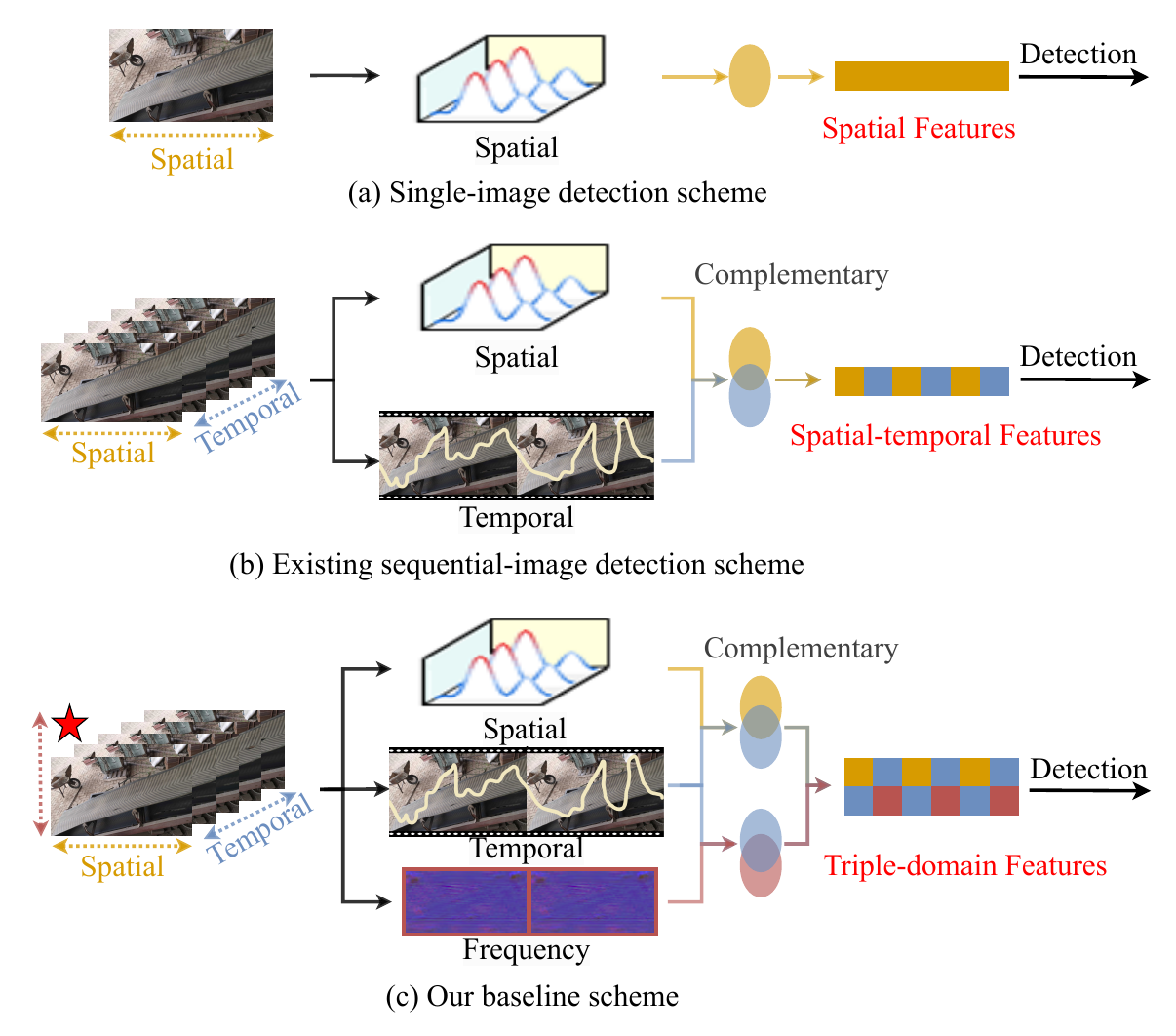}
	\caption{Comparison of three typical detection schemes for belt cracks: (a) single-image spatial scheme, (b) sequential-image spatio-temporal scheme, and (c) our sequential-image triple-domain scheme.}
	\label{fig:scheme}
\end{figure}

Currently, almost no learning-based methods have been proposed specifically for real-world belt crack detection. Existing approaches, such as PyramidFlow \citep{PyramidFlow}, RIND \citep{RINDNet} and AOST \citep{yangAOST2024}, mainly focus on general object detection or crack segmentation. These methods can be broadly grouped into two main categories.
One is the single-image methods relying only on spatial features \cite{CrackFormer}, as shown in Fig. \ref{fig:scheme}(a). The other is the sequential-image approaches with spatial-temporal fusion \citep{choi2024vid,li2024taptr}, still neglecting more feature domains, such as the frequency \cite{Duan2024}, 
as shown in Fig. \ref{fig:scheme}(b).

To validate the usefulness and effectiveness of our new datasets, 
we therefore propose a baseline framework by the triple-domain feature hierarchical learning with wavelet frequency. As shown in Fig. \ref{fig:scheme}(c), our method could capture spatial-temporal-frequency representations and hierarchically generate their fusion to enhance crack detection performance, surpassing conventional spatial–temporal frameworks.

In summary, the main contributions of this work can be summarized as follows:

(I) We construct and manually annotate two sequential-image belt crack detection datasets ($i.e.$, BeltCrack14ks and BeltCrack9kd). To the best of our knowledge, they are just the first pair of real-world sequential-image datasets for moving conveyor belts collected from an actual industrial production factory.

(II) We propose an initial baseline detection framework (BeltCrackDet) specifically for this pair of datasets. Beyond the traditional spatial or spatio-temporal feature modelling methods for cracks and general objects, it exploits triple-domain features (spatial, temporal and frequency) to enhance belt crack feature representation.

(III) We design and conduct extensive comparison and ablation studies to verify the availability and effectiveness of our new datasets. In the meanwhile, the effectiveness and advantages of the proposed baseline are validated.

\section{Related Work}
\label{sec2}
\subsection{Crack Detection Datasets}

Although industrial conveyor belts are widely used in modern manufacturing, there is still lack of real-world datasets for belt crack detection. This scarcity significantly hinders the development and application of learning-based methods. Existing crack datasets generally fall into two groups: general crack datasets and belt crack datasets.

(1) General Crack Datasets

General crack datasets primarily originate from infrastructure domains, such as roads, walls, and some even include artificially synthesized samples. For instance, Crack500 \citep{Yang2019} contains 500 pavement crack images captured by mobile devices, while CrackMap \citep{Katsamenis2023} comprises 114 samples for road crack segmentation. CrackSeg5k \cite{Chen2024CVPR} provides 2,000 nuclear power plant crack images as a domain-specific dataset. 
In addition, several composite datasets are built by aggregating smaller datasets to enhance diversity, e.g.,  CrackSeg9k \citep{Shreyas2023}, OMNICRACK30K \citep{Benz2024}, and TUT \cite{liu2024stair}.


(2) Belt Crack Datasets

Compared with general crack datasets, belt crack datasets are usually more difficult to collect, due to two main reasons: the confidentiality policies of industrial enterprises, and the harsh environment fo industrial belt running \citep{Peng2024}.


Existing belt crack datasets remain limited. MBTID \citep{Wang2022} comprises 1,708 images (372 torn, 491 scratched, and 845 intact). Part of the data is collected from real industrial sites, while the reset is generated in laboratory conditions using smoke generators to simulate realistic environments. CBCD \citep{Dwivedi2023} contains 1,562 crack samples from tunnel construction sites and experiment setups, supplemented by web-sourced images.

However, these datasets either come from non-industrial scenarios or are synthetically generated, lacking the complexity and diversity of real-world production environments. Moreover, they contain only single-frame images, which cannot capture the temporal evolution and dynamic formation of cracks during conveyor operation. 

\subsection{Traditional Crack Detection}

Early crack detection methods mainly relied on hand-crafted feature extraction, such as threshold-based segmentation \citep{SINKEVICIUS2015258}, edge detection \citep{John4767851}, and morphological operations \citep{Liu8296693}. While computationally efficient and easy to implement, these approaches show limited robustness in complex real-world environments.

Threshold-based methods depend heavily on predefined values, e.g., the Parzen window \citep{wang2008117}, and the histogram-based thresholding \citep{Tang6722269}.
This category of methods is often sensitive to noise, causing unstable results.

Furthermore, edge-based methods identify cracks by emphasizing local intensity gradients. Appleton \citep{Appleton20032513} introduced a circular shortest-path method for crack tracing, while  Ofir \citep{Ofir7780399} developed a multiscale edge detector for curved cracks under noisy conditions. 
Although effective in controlled scenarios, these approaches still  sensitive to noise and complex crack patterns, often resulting in incomplete localization.

Morphological operations, such as dilation and erosion, are commonly employed to refine crack maps. For example, Liu \citep{Liu8296693} combined morphology with visual features for concrete inspection, and Batool \citep{BATOOL2015642} presented Gabor filters with morphological processing for wrinkle detection. Nevertheless, these techniques rely heavily on hand-crafted features and threshold tuning, which limits their adaptability and generalization to real industrial environments.

\subsection{Learning-based Crack Detection}


(1) Single-image Detection

Since the introduction of convolutional neural networks (CNNs) for crack detection, substantial progress has been made \citep{JTFN,CrackNex,SimCrack}. For instance, SFIAN \citep{SFIAN} realizes real-time detection of irregular pavement cracks by multi-scale features and geometric cues. ADTM \citep{ADTM} improves segmentation by optimizing the binarization process, while DeepCrack \citep{Liu2019139} enables pixel-level segmentation inspired by edge detection. RIND \citep{RINDNet} models inter-class correlations to detect multiple crack types. 


Recently, Transformer-based models \citep{ashish2017attention} overcome this limitations by modeling global dependencies through self-attention. For example, CrackFormer \citep{CrackFormer} employs long-range context for fine-grained detection, VCVNet \citep{VCVNet} integrates Vision Transformer(ViT) with level-set theory for better boundary precision, and CCDFormer \citep{CCDFormer} combines global Transformer and local CNN branch. However, despite their superior representational capacity, Transformer often bring high computational and memory costs, making them less suitable for large-scale or real-time applications.


(2) Sequential-image Detection

Sequential-image methods leverage spatial-temporal correlation among consecutive frames to capture crack evolution and morphological changes.vid-TLDR \cite{choi2024vid} uses training-free token merging for lightweight video Transformers, preserving essential visual cues while improving efficiency. Similarly, TAPTR \cite{li2024taptr} formulates crack tracking as a point-query problem to enforce temporal consistency and reduce feature drift.

Despite leveraging temporal information, these methods still face challenges in real-world belt crack detection. In our task, single-image approaches often miss subtle or newly formed cracks due to motion blur and low contrast, while sequential methods remain sensitive to periodic belt deformation (e.g., shaft extrusion) and may confuse cracks with surface textures or stains. These challenges underscore the difficulty of crack detection in dynamic industrial environments and motivate the exploration of some robust frameworks.

\section{Our Proposed BeltCrack Datasets}
\label{sec3}
\subsection{Dataset Construction}

(1) Video Acquisition Equipment

Considering spatial constraints in industrial environments, data acquisition was conducted using identical smartphones for portability and ease of deployment. Videos were recorded at a 16:9 aspect ratio with resolutions of $3840\times2160$ and $1920\times1080$, at 30 FPS. To ensure stable recording, the smartphones were mounted on adjustable tripods to reduce motion blur. This setup provided a cost-effective yet reliable solution for fine-grained belt crack details under real industrial conditions.

(2) Data Acquisition Environment

The datasets were collected from a 55-meter heavy-duty rubber conveyor belt with regular anti-slip ridges and a smooth underside, forming noticeable textural variations. Naturally occurring cracks were observed, with longitudinal lengths ranging from 13–142 cm and transverse widths from 1–27 cm. During acquisition, the belt operated unloaded at speeds of 2–6 m/s, reflecting realistic industrial conditions.

Recordings were primarily captured under natural illumination, with auxiliary lighting used in low-light or shadowed conditions to ensure crack visibility. The data collection covered diverse weather scenarios (sunny, rainy, and snowy) and lighting conditions from daytime to evening. Both top-down and bottom-up views were recorded. Four independent sampling sessions were conducted over 3 consecutive days, each lasting approximately 6 hours, ensuring temporal and environmental diversity.

(3) Data Processing and Annotation

Each captured video was decomposed into sequential RGB frames. Based on crack saliency and density, the collected data were divided into two sub-datasets: BeltCrack14ks (strong visual saliency, dense crack regions) and BeltCrack9kd (weak saliency, sparse crack regions). Each original video was converted into a sequential dataset in COCO format, with frame-wise bounding box annotations. To ensure annotation quality, the labeling process included manual double-checking, cross-review among annotators, and spot validation. 

%
\subsection{Datasets Statistics}
Both datasets are partitioned strictly at the sequence level to ensure zero overlap. Specifically, the 29 sequences of BeltCrack14ks are split into train/validation/test sets by 18:5:6 sequences, yielding an approximate image-level ration of 6:2:2. Similarly, the 42 sequences of BeltCrack9kd are split into 33 train and 9 test sequences, maintaining an approximate 8:2 ratio. A comparative analysis shows that BeltCrack14ks has higher crack density (2.65 vs. 1.34 for BeltCrack9kd), and longer sequences (485.76 vs. 229.64 frames per sequence). These observations suggest that our BeltCrack9kd presents more challenging detection scenarios, with lower crack saliency and sparser spatial distributions, thereby increasing detection difficulty. To further quantify crack saliency, crack areas are calculated and categorized by size (tiny, small, medium, large, huge), as shown in Table \ref{table_ourdataset}.

\begin{table}
\centering
\renewcommand{\arraystretch}{1.8} 
\caption{Overall profiles. Sub.:sub-datasets, Img.: image number, Seq.: sequence number, CN: crack number, Img.$_{avg/seq}$: average image number per sequence, CN$_{avg/img}$: average crack number per image, Obs.: obscured. By region size (pixels), cracks are area-categorized into: tiny $(<100)$, small $(100 \to 1000)$, medium $(1000\to 10000)$, large $(10000\to 100000)$, and huge $(\ge 100000)$.}
\label{table_ourdataset}
\resizebox{\linewidth}{!}{
\begin{tabular}{l|cccccccccccc}
\hline\hline
\textbf{Datasets}              & \textbf{Sub.} & \textbf{Img.} & \textbf{Seq.} & \textbf{CN} & $\mathbf{Img._{avg/seq}}$  & $\mathbf{CN_{avg/img}}$    & \textbf{Obs.}    &\textbf{Tin.} & \textbf{Sma.} & \textbf{Med.} & \textbf{Lar.} & \textbf{Hug.}  
\\ \hline\hline
\multirow{3}{*}{BeltCrack14ks} & train           & 8773            & 18                 & 21475              & 487.39              & 2.79           & \textcolor{green}{\ding{52}}     & 5             & 588            & 12083           & 8755           & 44                                      \\
                               & val             & 2596            & 5                  & 7899               & 519.20               & 3.04          & \textcolor{green}{\ding{52}}      & 12            & 573            & 4700            & 2607           & 7                                       \\
                               & test            & 2718            & 6                  & 7932               & 453                 & 2.92        & \textcolor{green}{\ding{52}}          & 16            & 1117           & 3596            & 3196           & 7                           \\ 
                               
\hline
\multirow{2}{*}{BeltCrack9kd}  & train           & 7697            & 33                 & 10405              & 233.24              & 1.35             & \textcolor{green}{\ding{52}}       & -            & 16             & 2056            & 6330           & 2003                                \\
                               & test            & 1948            & 9                  & 2482               & 216.44              & 1.27        & \textcolor{green}{\ding{52}}         & -            & 2              & 362             & 767            & 1351                         \\ \hline
\end{tabular}}
\end{table}

\begin{figure}
	\centering
	\includegraphics[width=1.0\linewidth]{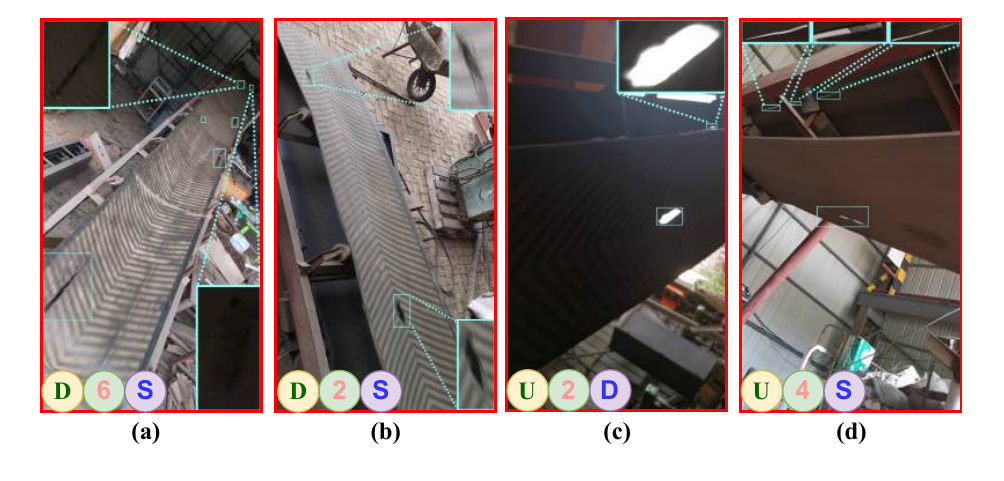}
	\caption{Samples in BeltCrack14ks with tags: camera  angle, crack count, lighting. (a) Down, 6 cracks and standard. (b) Down, 2 cracks and standard. (c) Up, 2 cracks and dim. (d) Up, 4 cracks and standard.}
	\label{fig:dataset1}
\end{figure}

\begin{figure}
	\centering
    \includegraphics[width=1.0\linewidth]{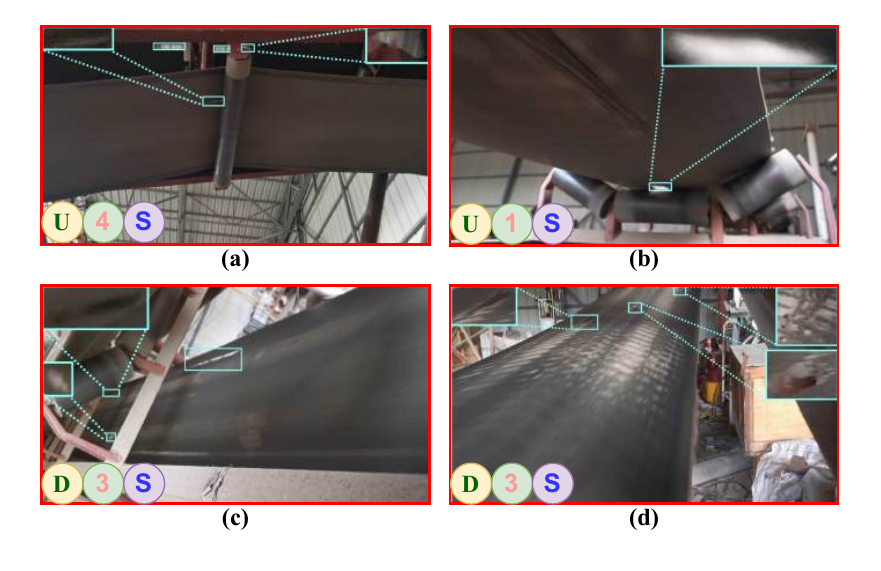}
    \caption{Samples in BeltCrack9kd with tags: camera  angle, crack count, lighting. (a) Up, 4 cracks and standard. (b) Up, 1 crack and standard. (c) Down, 3 cracks and standard. (d) Down, 3 cracks and standard.}
	\label{fig:dataset2}
\end{figure}

\begin{table*}[h]
\renewcommand{\arraystretch}{1.0} 
\caption{Crack dataset comparisons. Types: crack source; Ind.: industrial; Det.: detection; Seq.: sequential; Aug.: augmentation; BB: bounding box; N/I: max cracks/img; Res.: max resolution; Img.: image total.}
\label{table_com}
\resizebox{\linewidth}{!}{
\begin{tabular}{lcccccccccr}
\hline\hline
\textbf{Dataset} & \textbf{Year} & \textbf{Types}  & \textbf{Ind.}       & \textbf{Det.} & \textbf{Seq.} & \textbf{Aug.} & \textbf{BB} & \textbf{N/I} & \textbf{Res.}                                         & \textbf{Img.} \\ 
\hline\hline
Crack500\citep{Yang2019}        & 2019          & pavement  & \textcolor{red}{\ding{56}} & \textcolor{red}{\ding{56}}            & \textcolor{red}{\ding{56}}            & \textcolor{green}{\ding{52}}           & \textcolor{red}{\ding{56}}          & 1            & $2000\times 1500$ & 500           \\
CrackSeg9k\citep{Shreyas2023}       & 2022          & farraginous  &  \textcolor{red}{\ding{56}}   & \textcolor{red}{\ding{56}}            & \textcolor{red}{\ding{56}}            & \textcolor{green}{\ding{52}}           & \textcolor{red}{\ding{56}}          & 1            & $400 \times 400$                                              & 9,255         \\
MBTID\citep{Wang2022}            & 2022          & belt    &\textcolor{green}{\ding{52}}       & \textcolor{red}{\ding{56}}            & \textcolor{red}{\ding{56}}            & \textcolor{green}{\ding{52}}           & \textcolor{red}{\ding{56}}          & 1            & $416\times 416$                                              & 1,708         \\
CBCD\citep{Dwivedi2023}             & 2023          & belt  &\textcolor{green}{\ding{52}}     & \textcolor{red}{\ding{56}}            & \textcolor{red}{\ding{56}}            & \textcolor{red}{\ding{56}}            & \textcolor{red}{\ding{56}}          & 1            & $608\times 608$                      & 1,562         \\
CrackMap\citep{Katsamenis2023}        & 2023          & road  &\textcolor{red}{\ding{56}}                       & \textcolor{red}{\ding{56}}            & \textcolor{red}{\ding{56}}            & \textcolor{green}{\ding{52}}           & \textcolor{red}{\ding{56}}          & 1            & $5184\times3888$                                            & 114           \\

CrackSeg5k\cite{Chen2024CVPR}    &2024 & wall  &\textcolor{green}{\ding{52}}                       & \textcolor{red}{\ding{56}}            & \textcolor{red}{\ding{56}}            & \textcolor{green}{\ding{52}}           & \textcolor{red}{\ding{56}}          & 1            & $7360\times 4912$                                            & 2,000        \\

OMNICRACK30K\citep{Benz2024}     & 2024          & farraginous  &\textcolor{red}{\ding{56}}                 & \textcolor{red}{\ding{56}}            & \textcolor{red}{\ding{56}}            & \textcolor{green}{\ding{52}}           & \textcolor{red}{\ding{56}}          & 1            & $4608\times 5184$                                            & 30,017        \\

TUT\cite{liu2024stair}  &2025 &farraginous &\textcolor{red}{\ding{56}}                       & \textcolor{red}{\ding{56}}            & \textcolor{red}{\ding{56}} &\textcolor{green}{\ding{52}}     & \textcolor{red}{\ding{56}}  &1 & $640 \times 640$ & 1,408  \\

\textbf{BeltCrack14ks(Ours)}    & 2025       & belt &\textcolor{green}{\ding{52}}                        & \textcolor{green}{\ding{52}}           & \textcolor{green}{\ding{52}}           & \textcolor{red}{\ding{56}}            & \textcolor{green}{\ding{52}}         & 10           &$ 2160\times3840$                                            & 14,087        \\
\textbf{BeltCrack9kd(Ours)}     & 2025       & belt &\textcolor{green}{\ding{52}}                        & \textcolor{green}{\ding{52}}           & \textcolor{green}{\ding{52}}           & \textcolor{red}{\ding{56}}            & \textcolor{green}{\ding{52}}         & 7            &$ 2160\times3840$               & 9,645         \\ \hline
\end{tabular}}
\end{table*}

Moreover, the two datasets were collected under four diverse conditions: (i) multi-view imaging (e.g., different camera angles), (ii) varying illumination (morning, noon, dusk, with/without auxiliary lighting), (iii) multiple weather scenarios (sunny, snowy, overcast), and (iv) varying conveyor speeds (2-6 m/s). These environmental variations provide substantial diversity to support model generalization. As illustrated by representative samples from each dataset in Fig. \ref{fig:dataset1} and Fig. \ref{fig:dataset2}, our datasets exhibit broader environmental diversity than existing datasets summarized in Table \ref{table_com}.

\begin{figure*}
    \centering
    \includegraphics[width=1.0\textwidth]{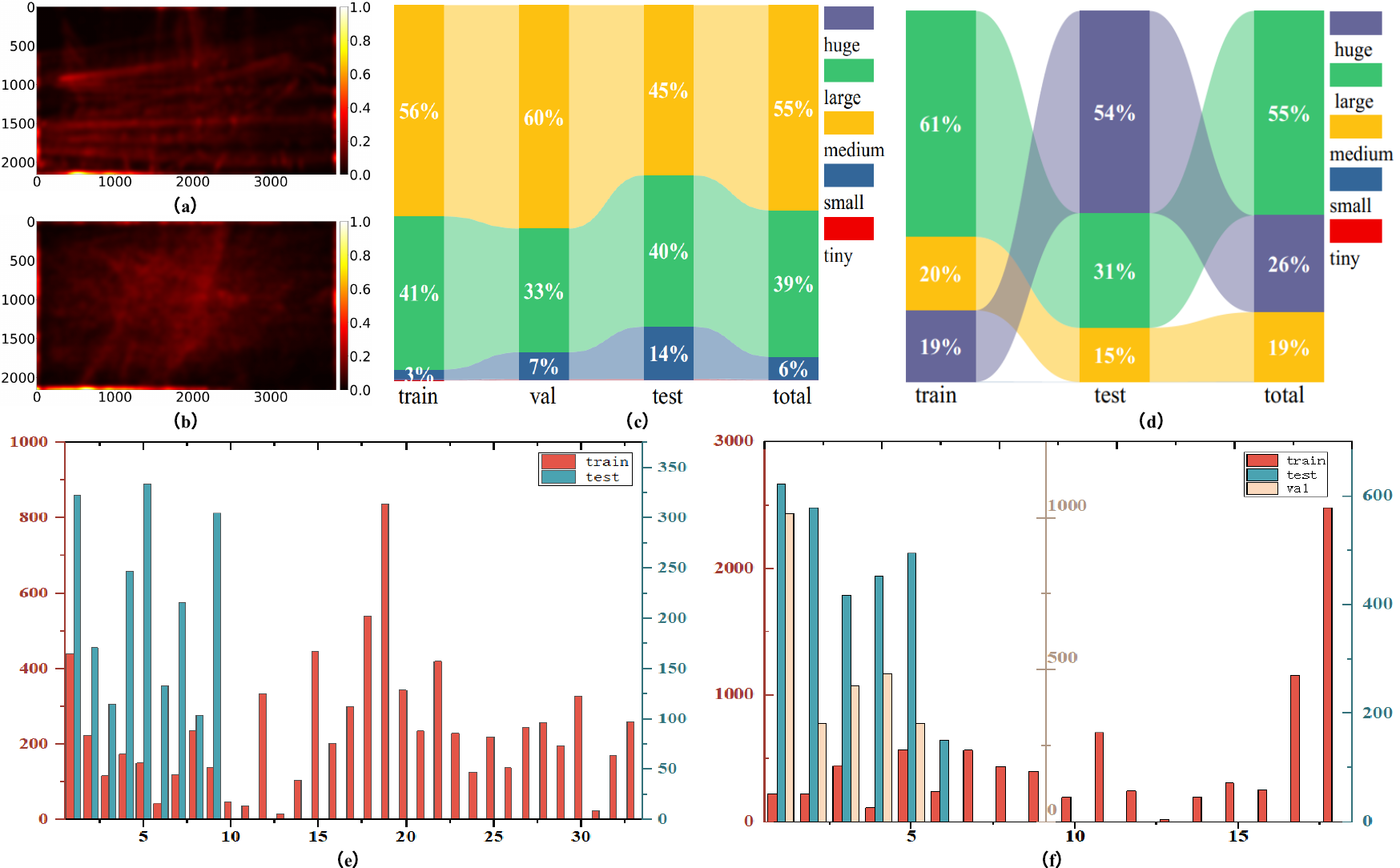}
    \caption{Visualization of crack characteristics in BeltCrack14ks and BeltCrack9kd: (a-b) spatial localization, (c-d) pixel-level area, and (e-f) temporal sequence length.}
    
    \label{fig:visual_dataset}
\end{figure*}

As quantified in Fig. \ref{fig:visual_dataset}(a) and \ref{fig:visual_dataset}(b), spatial distribution analysis indicates that BeltCrack9kd has more dispersed crack patterns, suggesting its potential value for studying sparsely distributed cracks. Cracks are further categorized by area, with the size criteria summarized in Table \ref{table_ourdataset}. As illustrated in Fig. \ref{fig:visual_dataset}(c) and \ref{fig:visual_dataset}(d), both datasets exhibit that crack areas approximately follow normal distributions with long-tailed behavior. These statistics indicate that the samples in our datasets are sufficient, diverse and statistically representative, which facilitates comprehensive exploration of crack characteristics. Moreover, sequence length distributions are presented in Fig. \ref{fig:visual_dataset}(e) and \ref{fig:visual_dataset}(f). Both datasets show uneven sequence lengths, mainly due to belt motion, occlusions and view limitations during acquisition. This variability realistically reflects industrial conditions, while the larger variation in BeltCrack9kd highlight higher complexity and benchmark value for industrial inspection tasks.

\subsection{Superiority over Existing Datasets}
The proposed datasets could advance learning-based industrial belt crack detection and offer several clear advantages, as summarized in Table \ref{table_com}.

(1) Larger-scale Datasets

We provide over 20,000 high-resolution sequential images together with dense, frame-wise crack annotations, enabling richer supervision and more reliable statistics. Compared with existing datasets that are typically limited in image quantity, spatial coverage, or annotation density, our datasets offer a larger scale and finer granularity, enabling more comprehensive model training and evaluation.

(2) Sequence-based, Not Image-based

To the best of our knowledge, these are the first belt-crack benchmark datasets collected for industrial time-series scenarios. Unlike static image collections, both datasets capture crack evolution across conveyor operating cycles, enabling spatiotemporal modeling beyond conventional single-image detection.

(3) Real-world Industrial, Not Synthetic

All data are collected from real industrial production environments rather than simulated sources. In contrast, datasets such as MBTID \citep{Wang2022} and CBCD \citep{Dwivedi2023} include artificially generated samples based on small-scale laboratory setups.

\section{Proposed Baseline for Crack Detection}
\label{sec4}
\subsection{Motivation}
Currently, most data-driven crack detection methods are designed for single-frame \citep{Dwivedi2023} analysis, with only a few extending to multi-frame \citep{Kuang2025CVPR} inputs to capture limited temporal cues. However, these approaches are primarily designed for static infrastructure scenarios and are not well suited to industrial belt environments, where continuous motion, lighting fluctuations, and texture variations are common. 

To address these limitations, we propose \textbf{BeltCrackDet}, a baseline specially proposed for sequential industrial belt crack detection. It adopts a triple-domain hierarchical fusion strategy that integrates spatial, temporal, and frequency features to enhance robustness under complex, dynamic industrial conditions.

\subsection{Overall Architecture}

\begin{figure*}
    \centering
    \includegraphics[width=1.0\textwidth]{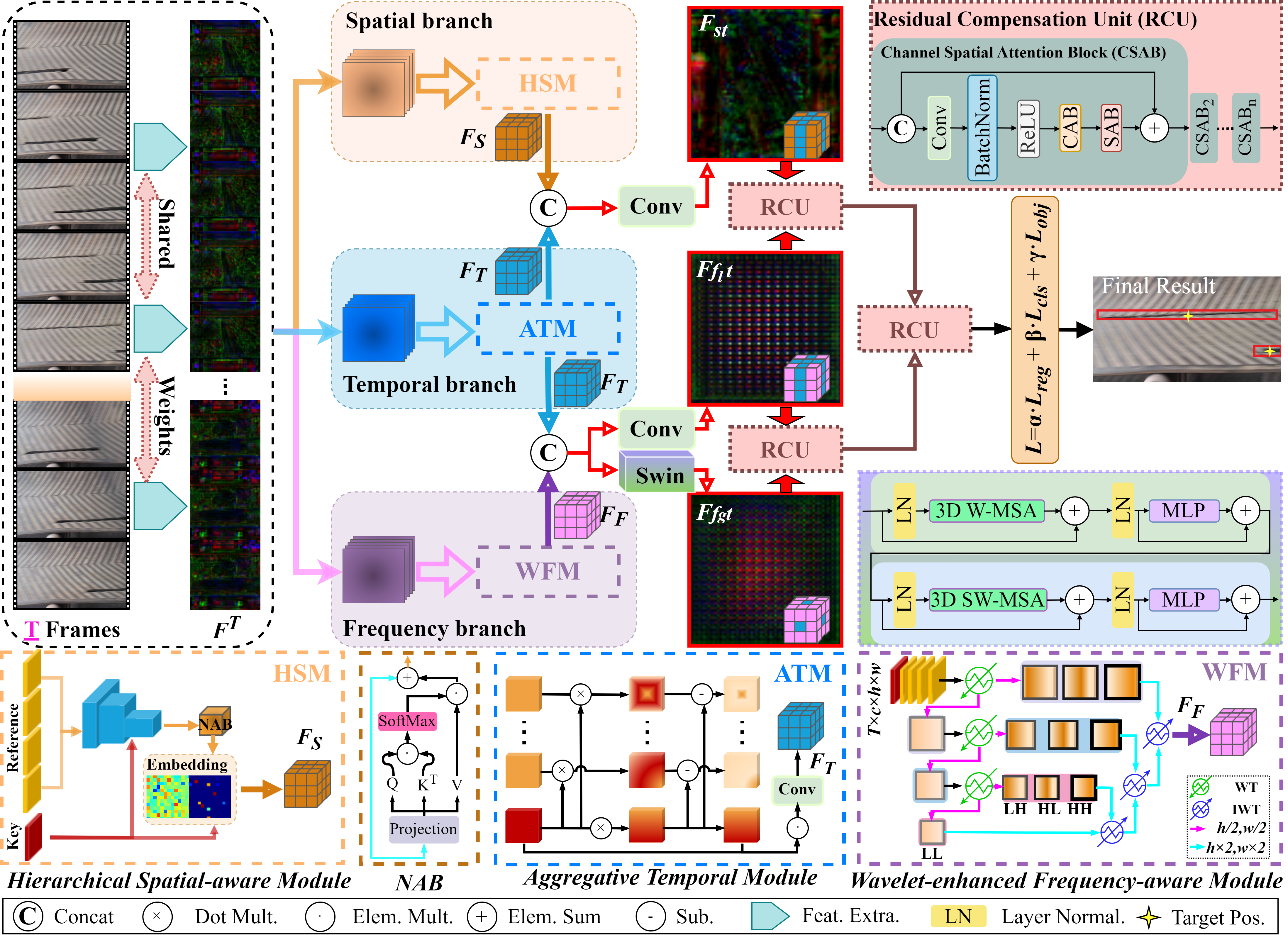}
    \caption{Overview of our BeltCrackDet, with three feature extraction branches: the spatial-domain features, the temporal-domain features, and the frequency-domain features.   
    }
    \label{fig:arch}
\end{figure*}

In detail, $T$ denotes the number of frames sampled from the same sequence, and $H$ and $W$ represent the image height and width, respectively. We adopt CSPDarknet \citep{Ge2021YOLOXEY} as the backbone for feature extraction, though any modern backbone can be substituted. Each frame in $ \boldsymbol{I}^T =\left \{\boldsymbol{I_1},\boldsymbol{I_2},...,\boldsymbol{I_t} \right \}$ is processed by the same backbone with shared weights. Here, $\boldsymbol{I_t}$ denotes the key-frame, while the preceding frames provide dynamic temporal context. This yields a multi-frame feature set $\textbf{\textit{F}}^{\textbf{\textit{T}}}= \left \{\boldsymbol{F_1} ,\boldsymbol{F_2},...,\boldsymbol{F_t} \right \}\in \mathbb{R}^{T\times c\times h\times w}$, where $c$, $h$, $w$ denote the channel, height, and width dimensions, respectively.

To enrich feature representations, we propose a tri-path network architecture for cross-domain representation learning through three specialized modules: \textit{Hierarchical Spatial-aware Module} (HSM), \textit{Aggregative Temporal Module} (ATM), and \textit{Wavelet-enhanced Frequency-aware Module} (WFM). Additionally, a \textit{Residual Compensation Unit} (RCU) is introduced to mitigate domain discrepancies and improve cross-domain feature fusion. The enhanced frequency-spatial-temporal features $\boldsymbol{F_{fst}}$ are subsequently fed into the detection head for crack localization and classification.

Specifically, the HSM takes the multi-frame features $\boldsymbol{F^T}$ generated by the backbone as input to derive refined spatial representations $\boldsymbol{F_S}$. Meanwhile, the ATM is designed to extract and enhance the spatial-temporal consistency $\boldsymbol{F_T}$ by aligning multi-frame inputs. To jointly capture location and motion information of crack regions, spatial features $\boldsymbol{F_S}$ and temporal features $\boldsymbol{F_T}$ are concatenated and fused through a convolutional operation, forming spatial-temporal features $\boldsymbol{F_{st}}$, as defined by
\begin{equation}
\label{eq1}
\boldsymbol{F_{st}}  = Conv\left ( Concat\left [ \boldsymbol{F_S} ,\boldsymbol{F_T} \right ]  \right ) ,
\end{equation}
where  $Conv$ denotes a $3 \times 3$ convolution with batch normalization (BN) and SiLU activation function.

In parallel, the WFM extracts frequency-domain features $\boldsymbol{F_F}$ by transforming the multi-frame features $\boldsymbol{F^T}$. This frequency-domain analysis captures both motion cues and structural regularities, thereby enhancing the crack features representations.
\begin{equation}
    \label{eq2}
    \begin{cases}
    \boldsymbol{F_{f_{l}t}} = Conv\left ( Concat\left[\boldsymbol{F_F},\boldsymbol{F_T}\right ]  \right )  ,
     \\
    \boldsymbol{F_{f_{g}t}} = Conv\left ( Swin\left (  Concat\left [\boldsymbol{F_F}, \boldsymbol{F_T}\right ]  \right )  \right ),
\end{cases}
\end{equation}
where $\boldsymbol{F_{f_{l}t}}$ and $\boldsymbol{F_{f_{g}t}}$ denote the local and global fused features, respectively. Moreover, \textit{Swin(.)} representing a Swin-Transformer-based block.

Additionally, three RCUs are introduced to facilitate cross-domain feature interaction and complementary representation learning. Specifically, the first two RCUs refine the local ($\boldsymbol{F_{f_{l}t}}$) and global ($\boldsymbol{F_{f_{g}t}}$) frequency–temporal representations through integration with the spatial–temporal features $\boldsymbol{F_{st}}$. These refined outputs are then fused by the final RCU to generate compensatory frequency-spatial-temporal features $\boldsymbol{F_{fst}}$, as expressed by
\begin{equation}
    \label{rcu_triple}
    \boldsymbol{F_{fst}} = RCU_{3}\!\left(
    RCU_{1}\!\left(\boldsymbol{F_{f_{l}t}}, \boldsymbol{F_{st}}\right),
    RCU_{2}\!\left(\boldsymbol{F_{f_{g}t}}, \boldsymbol{F_{st}}\right)
\right).
\end{equation}
This hierarchical compensation process effectively reduces cross-domain feature discrepancies and enhances the robustness of crack representation learning. Finally, crack detection is then obtained using the decoupled head of YOLOX \citep{Ge2021YOLOXEY}, followed by non-maximal suppression.

\subsection{Hierarchical Spatial-aware Module}

Inter-frame spatial context is crucial for industrial conveyor belt crack detection, as it captures spatial dependencies that help resolve ambiguities caused by occlusion, illumination variations, and motion blur. To this end, the HSM is designed to capture and aggregate spatial relationships across consecutive frames, as illustrated in Fig. \ref{fig:arch}. By hierarchically combining local fusion, a \textit{Non-local Attention Block} (NAB), and a \textit{Memory Enhancement Unit} (MEU), the HSM produces consistent spatial representations, capturing long-range dependencies and providing a solid foundation for subsequent temporal and frequency-domain modeling.

Specifically, inspired by the visual tracking system \citep{Duan2024}, the HSM first performs local cross-frame fusion to refine the key-frame features $\boldsymbol{F_t}$ by integrating reference information from preceding frames $\left \{\boldsymbol{F_1} ,\boldsymbol{F_2},...,\boldsymbol{F_{t-1}} \right \}$. All reference features are concatenated and passed through a convolutional gating unit to generate an adaptive spatial map, which is used to dynamically refines the key-frame features. The resulting features are then processed by additional convolutional block to obtain the spatially fused representation, by
\begin{equation}
    \label{eq4}
    \begin{cases}
        \boldsymbol{ \widehat{F_{t}}}  &= \sigma \left ( Conv\left ( Concat\left [ \boldsymbol{F_{1}} ,...,\boldsymbol{F_{t-1}} \right ]  \right )  \right )\odot \boldsymbol{F_{t}} , \\
         
        \boldsymbol{F_{l}}  &= Conv\left ( Concat\left [ \widehat{\boldsymbol{F_{t}}} ,\boldsymbol{F_{t}}    \right ]  \right ) ,
    \end{cases}
\end{equation}
where $\sigma \left ( x\right)=\frac{1}{1+e^{-x}}$ represents the sigmoid activation,  and $\odot$ denotes element-wise multiplication. Here, $\boldsymbol{\widehat{F_t}}$ is the adaptively modulated feature map, while $\boldsymbol{F_{l}}$ indicates local inter-frame spatial dependencies.

While the above step models local dependencies, the NAB is designed to capture long-range dependencies that conventional convolutional layers struggle to model. Given the fused feature $ \boldsymbol{F_{l}}$ as input, it models correlations between spatial positions to enhance global structural awareness. Specifically, it first generates query, key, and value feature maps through linear projections and then computes the attention map as:
\begin{equation}
    \label{eq5}
    \begin{cases}
    ( \boldsymbol{Q}, \boldsymbol{K}, \boldsymbol{V}) = \mathrm{MatMul}(\boldsymbol{F_l}, (\boldsymbol{W^Q}, \boldsymbol{W^K}, \boldsymbol{W^V})), \\
    \boldsymbol{F_g} = \mathrm{Softmax}\!\left(\frac{\boldsymbol{QK^{T}}}{\sqrt{d_k}}\right)V + \boldsymbol{F_l},
    \end{cases}
\end{equation}
where $\boldsymbol{W^Q},\boldsymbol{W^K}$, and $\boldsymbol{W^V}$ are learnable matrices, and $\sqrt{d_{k}}$ is a normalization constant. The resulting feature $\boldsymbol{F_{g}}$ captures local-global spatial relations. Through the NAB, spatial positions across the feature map can directly interact, enabling the network to model long-range spatial dependencies beyond the limitations of local receptive fields.

\begin{figure}
    \centering
    \includegraphics[width=1.0\linewidth]{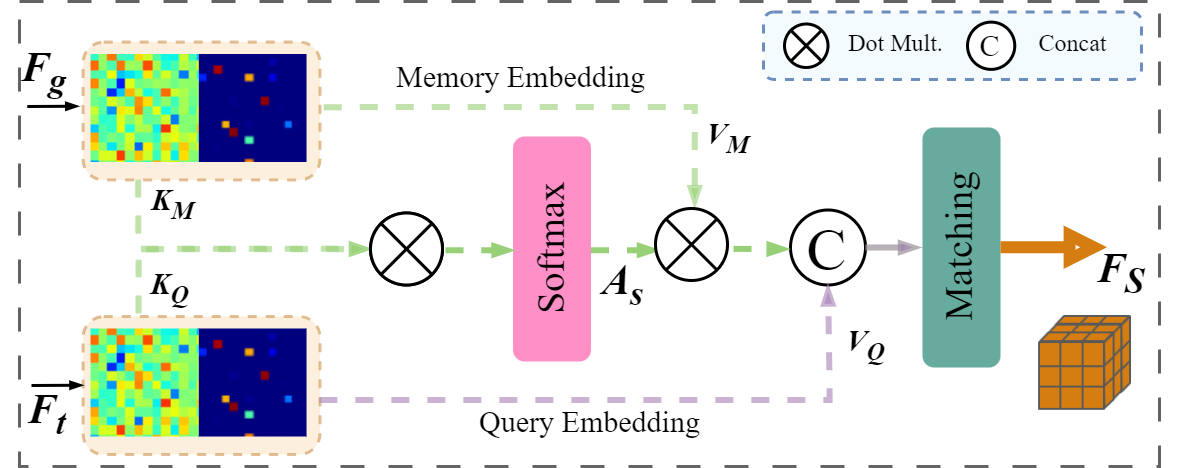}
    \caption{Overflow of MEU. It establishes the connection between key-frame and reference frames.}
    \label{fig:meu}
\end{figure}

To further enhance the temporal consistency of spatial representations, the MEU (see Fig. \ref{fig:meu}) adopts a memory-based query–key mechanism to model spatial-temporal dependencies across frames. The local-global features $\boldsymbol{F_{g}}$ obtained from NAB are stored in memory, while the key-frame features $\boldsymbol{F_t}$ act as queries. The corresponding key–value pairs are computed as:
\begin{equation}
    \label{eq6}
    \begin{cases}
    (\boldsymbol{K_Q}, \boldsymbol{V_Q}) = (\mathrm{Conv}_Q^K(\boldsymbol{F_t}),\, \mathrm{Conv}_Q^V(\boldsymbol{F_t})), \\
   (\boldsymbol{K_M}, \boldsymbol{V_M}) = (\mathrm{Conv}_M^K(\boldsymbol{F_g}),\, \mathrm{Conv}_M^V(\boldsymbol{F_g})),
    \end{cases}
\end{equation}
where $\mathrm{Conv}_Q^K$ and $\mathrm{Conv}_Q^V$ generate the key and value maps for the query branch, while $\mathrm{Conv}_M^K$ and $\mathrm{Conv}_M^V$ produce the corresponding maps for the memory features.

The similarity between the key-frame features and memory features is computed using a normalized attention mechanism:
\begin{equation}
    \label{MEU_FT}
    \begin{cases}
    \boldsymbol{A_s} = \text{Softmax}(\boldsymbol{K_M} \otimes \boldsymbol{K_Q}), \\
    \boldsymbol{F_S} = \text{Matching}(\text{Concat}[\boldsymbol{A_s} \otimes \boldsymbol{V_M}, \boldsymbol{V_Q}]),
    \end{cases}
\end{equation}
where $\boldsymbol{A_s}$ is the softmax-normalized similarity matrix computed from $\boldsymbol{K_M}$ and $\boldsymbol{K_Q}$; The Matching denotes a $1\times1$ convolution used to fuse contextual relations; and $\boldsymbol{F_S}$ represents the enhanced local–global spatial features.

Through this design, the MEU dynamically aligns current key-frame representations with memory from adjacent frames, helping the model maintain boundary continuity and contextual coherence during crack detection.

\subsection{Aggregative Temporal Module}
\begin{figure}
    \centering
    \includegraphics[width=1.0\linewidth]{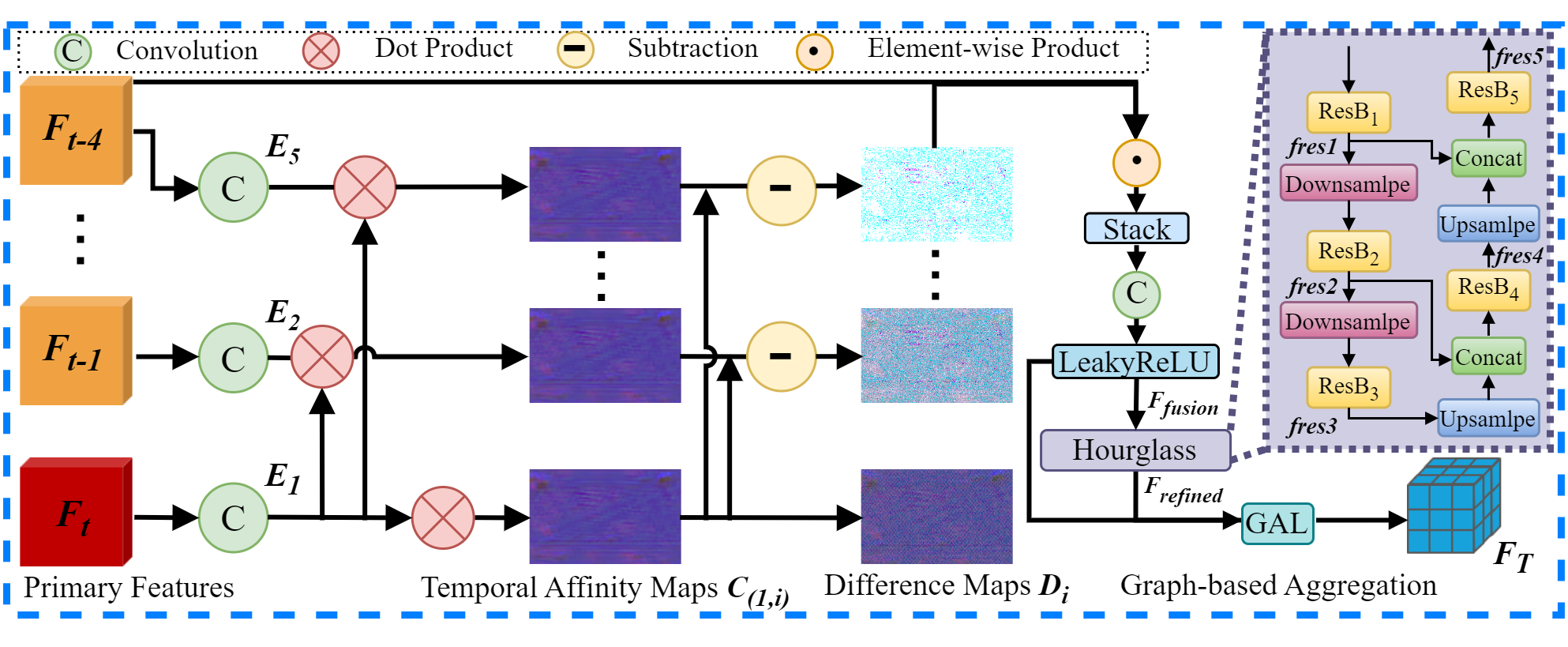}
    \caption{Overview of the proposed ATM, with time window $T=5$. It aggregates motion-consistent temporal features and enhances spatial topology for robust crack representation.}
    \label{fig:ATM}
\end{figure}

Extracting motion characteristics of belt cracks is challenging in industrial environments due to the high-speed motion of conveyor belts. Crack regions are often subject to deformation and interference from objects, making it difficult to capture stable temporal cues across consecutive frames. To address these challenges, our ATM introduces a dual-stage architecture to capture motion-consistent and structurally coherent temporal representations, as shown in Fig. \ref{fig:ATM}. 

In detail, each feature in $\boldsymbol{F^{T}}$ is fed into the ATM to model inter-frame motion. As shown in  Eq. (\ref{affinity}), each feature $\boldsymbol{F_i}$ is first embedded through a convolutional layer to obtain $\boldsymbol{E_i}$. Then, temporal affinity maps $\boldsymbol{C_{(i,j)}}$ are computed to measure pixel-wise correlations between each reference embedding $\boldsymbol{E_i}$ and the key-frame embedding $\boldsymbol{E_1}$.
\begin{equation}
\label{affinity}        
    \boldsymbol{E_i}=Conv(\boldsymbol{F_i}),  \quad 
    \boldsymbol{C_{(1,i)}} = \sum_{c=1}^{C} \left(\boldsymbol{E_1^{c}}  \odot  \boldsymbol{E_i^{c}}\right), \quad i\ne 1, 
\end{equation}
where $\odot$ denotes element-wise multiplication along the channel dimension, and the summation aggregates responses over all $C$ channels. This operation captures motion-consistent regions across frames, forming the basis for subsequent temporal refinement.

To emphasize dynamic crack variations between neighboring frames, temporal difference maps $\boldsymbol{D_i}$ are computed based on the Euclidean distance between each reference frame and the key frame. These maps are then used to derive a motion-weighted fusion representation $\boldsymbol{F_{fusion}}$, formed by concatenating motion-weighted features of all frames, followed by a $1\times1$ convolution and a nonlinear activation. A multi-scale Hourglass network further refines $\boldsymbol{F_{fusion}}$ through two stages of downsampling and upsampling with residual blocks and skip connections. This design captures multi-scale contextual cues and improves crack boundary localization, as formulated in Eq. (\ref{difference}):
\begin{equation}
\label{difference}
    \begin{cases}
        \boldsymbol{D_{1}} =\boldsymbol{C_{(1,1)}},        \boldsymbol{D_{i}}  =| \boldsymbol{C_{(1,1)}}-\boldsymbol{C_{(1,i)}} |, i\ne 1;  \\
        \boldsymbol{\widehat{E_i}}  = \sigma {(\boldsymbol{D_{i}})} \odot \boldsymbol{E_{i}}; \\
        \boldsymbol{F_{fusion}}=\phi\!\left(Conv\!\left(Stack\!\left([\boldsymbol{\widehat{E}_{i}]_{i=1}^{T}}\right)\right)\right), \\ \boldsymbol{F_{refined}}=Hourglass(\boldsymbol{F_{fusion}}),
    \end{cases}
\end{equation}
where $T=5$, $\sigma(\cdot)$ is the sigmoid function for motion-consistency weighting, and $\phi(\cdot)$ denotes the Leaky ReLU activation. The \textit{Hourglass} module refines motion features in a multi-scale manner, as illustrated in Fig. \ref{fig:ATM}.

Although the Hourglass network improves local motion perception and boundary localization, its receptive field remains limited and it lacks explicit modeling of long-range spatial–temporal dependencies. To address this limitation, we introduce a graph-based aggregation layer (GAL) that enables directional message passing and structural feature aggregation.

We first obtain the initial vertex feature $\boldsymbol{V}$ by integrating $\boldsymbol{F_{fusion}}$ and $\boldsymbol{F_{refined}}$ through a $1\times1$ convolution followed by Leaky ReLU activation. Each spatial location in $\boldsymbol{V}$ is treated as a vertex, and four-connected neighbors ($\uparrow, \downarrow, \leftarrow, \rightarrow$) are regarded as directed edges. The corresponding edge features $\boldsymbol{E}_k$ are computed via element-wise multiplication between each vertex and its spatially shifted neighbors $\mathcal{S}_k(\boldsymbol{V})$, where $\mathcal{S}_k(\cdot)$ denotes the spatial shift operator in direction $k$, as shown in Eq. (\ref{gal1}).
\begin{equation}
    \label{gal1}
    \begin{cases}
    \boldsymbol{V} = \phi(\mathrm{Conv}_{1\times1}([\boldsymbol{F_{fusion}}, \boldsymbol{F_{refined}}])), 
    \\  
    \boldsymbol{E}_k = \boldsymbol{V} \odot \mathcal{S}_k(\boldsymbol{V}), 
    \quad 
    k \in \{\uparrow, \downarrow, \leftarrow, \rightarrow\}.
    \end{cases}
\end{equation}
Next, the aggregated edge feature $\boldsymbol{A}$ is obtained by a linear projection followed by batch normalization and ReLU. The vertex feature $\boldsymbol{U_v}$ is updated by concatenating $\boldsymbol{A}$ and $\boldsymbol{V}$ and passing them through another linear mapping. The edge feature $\boldsymbol{R}$ is further refined via directional transformation and weighted aggregation. Finally, the updated vertex and edge representations are combined via element-wise modulation and a $1\times1$ convolution to produce the final output feature $\boldsymbol{F_T}$:
\begin{equation}
    \label{gal2}
    \begin{cases}
    \boldsymbol{A} = \rho(\mathrm{BN}  (\boldsymbol{E}\boldsymbol{W}_e)),  \quad
    \boldsymbol{U_v} = \rho(\mathrm{BN}(\boldsymbol{W}_v[\boldsymbol{V};\boldsymbol{A}])), \\
    \boldsymbol{R} = \rho\!\left(\mathrm{BN}\!\left(\sum_{k=1}^{4}\alpha_k \boldsymbol{W}_r[\boldsymbol{V};\boldsymbol{E}_k]\right)\right), \\ 
    \boldsymbol{F_T} = \phi(\mathrm{Conv}_{1\times1}([\boldsymbol{V}, \boldsymbol{U_v} \odot \boldsymbol{R}])),
    \end{cases}
\end{equation}
where $\rho(\cdot)$ and $\phi(\cdot)$ denote the ReLU and Leaky ReLU activations, respectively. The matrices $\boldsymbol{W}_e$, $\boldsymbol{W}_v$, and $\boldsymbol{W}_r$ are learnable. $\alpha_k$ represents the directional weight for the $k$-th neighbor, and $\odot$ indicates element-wise multiplication.

\subsection{Wavelet-enhanced Frequency-aware Module}

Accurately modeling frequency-domain characteristics of cracks under dynamic and noisy backgrounds is essential for robust detection. To this end, we propose the WFM, inspired by wavelet theory \citep{Yan2025CVPR}. Traditional convolutional layers operate in the spatial domain, which limits their ability to distinguish high-frequency crack edges from low-frequency background textures. In contrast, wavelet transform decomposes features into multiple frequency subbands, allowing fine-grained separation and enhancement of texture and edge information critical for belt crack detection \citep{Lu2023ICCV}.

\begin{figure}
    \centering
    \includegraphics[width=1.0\columnwidth]{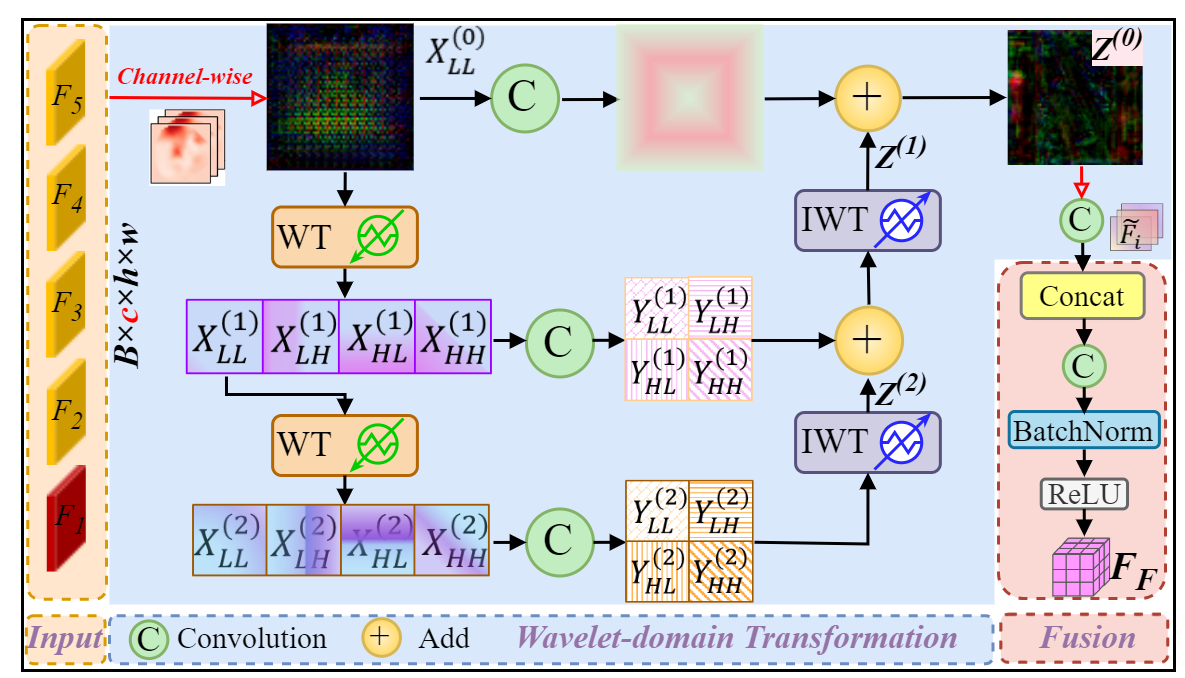}
    \caption{WFM with two-level decomposition. It generates $F_F$ by fusing $T=5$ frames through channel-wise wavelet decomposition, frequency-domain convolution, and hierarchical IWT reconstruction.}
    \label{fig:WFM}
\end{figure}

As illustrated in Fig. \ref{fig:WFM}, the WFM processes each feature map  in $\boldsymbol{F^T}$ through multi-level wavelet decomposition. Specifically, the decomposition is performed hierarchically by recursively applying the \textit{Wavelet Transform} (WT) to the low-frequency component at each level $j$, as defined in Eq. (\ref{WT-cas}):
\begin{equation}
    \label{WT-cas}
\boldsymbol{X_{LL}^{(j)}},\boldsymbol{X_{LH}^{(j)}},\boldsymbol{X_{HL}^{(j)}},\boldsymbol{X_{HH}^{(j)}} = WT(\boldsymbol{X_{LL}^{(j-1)}}),
\end{equation}
where $\boldsymbol{X_{LL}^{(0)}}$ denotes the input feature, $\boldsymbol{X_{LL}^{(j)}}$ is the low-frequency component at level $j$, and $\boldsymbol{X_{LH}^{(j)}}$ , $\boldsymbol{X_{HL}^{(j)}}$, and $\boldsymbol{X_{HH}^{(j)}}$ represent the horizontal, vertical, and diagonal high-frequency subbands, respectively.

To extract discriminative frequency cues, the input feature map is first decomposed into multiple frequency subbands through WT. A small-kernel depth-wise convolution is then independently applied to each subband to enhance local frequency responses, followed by the \textit{Inverse Wavelet Transform} (IWT) for reconstruction, as expressed by
\begin{equation}
    \label{WT-out}
    \begin{cases}
    \boldsymbol{Y} = IWT(Conv(\boldsymbol{W},WT(\boldsymbol{X}))), \\
    \boldsymbol{Y_{LL}^{(j)}},\boldsymbol{Y_{LH}^{(j)}},\boldsymbol{Y_{HL}^{(j)}},\boldsymbol{Y_{HH}^{(j)}}= Conv\left ( \boldsymbol{W^{(j)}},\left ( \boldsymbol{X_{LL}^{(j)}},\boldsymbol{X_{H}^{(j)}} \right )  \right ) ,
    \end{cases}
\end{equation}
where $\boldsymbol{X}$ is the input feature map, and $\boldsymbol{W}$ is the learnable depthwise convolution kernel for frequency subbands. At level $j$, $\boldsymbol{W^{(j)}}$ denotes the kernel for that decomposition stage, while $\boldsymbol{X_{H}^{(j)}}$ collectively refers to the three high-frequency subbands.

Through hierarchical reconstruction, low- and high-frequency components are progressively fused in a top-down manner. At level $j$, the reconstructed low-frequency feature $\boldsymbol{Y_{LL}^{(j)}}$ is combined with the aggregated high-frequency components $\boldsymbol{Z^{(j+1)}}$ obtained from the next level, yielding the full-resolution output via IWT:
\begin{equation}
    \label{IWT-level}
    \begin{cases}
        \boldsymbol{Y_{n}^{(j)}} = \boldsymbol{Y_{LL}^{(j)}} +  \boldsymbol{Z^{(j+1)}} , \\
        \boldsymbol{Z^{(j)}} =IWT(\boldsymbol{Y_{n}^{(j)}},\boldsymbol{Y_{LH}^{(j)}},\boldsymbol{Y_{HL}^{(j)}},\boldsymbol{Y_{HH}^{(j)}}),
    \end{cases}
\end{equation}
where $\boldsymbol{Z^{(j+1)}} =0$ at the highest decomposition level, $j$ denotes the decomposition levels, and $\boldsymbol{Z^{(j)}}$ represents the aggregated reconstruction results from level $j$ down to 0.

Finally, the reconstructed features $\boldsymbol{Z^{\left (0\right )}}$ from all frames are aggregated to obtain a hybrid frequency-domain representation that preserves strong frequency discriminability. Applying the wavelet-based convolutional transformation across $T$ consecutive frames yields overall frequency-domain representation $\boldsymbol{F_{F}}$, formulated as:
\begin{equation}
    \label{wfm_res}
    \begin{cases}
    \widetilde{\boldsymbol{F}}_{i} = \mathcal{W}(\boldsymbol{F}_{i}), \\
    \boldsymbol{F_F} = SiLU\!\big(BN(Conv(Concat\big[\widetilde{\boldsymbol{F}}_{1}, \ldots, \widetilde{\boldsymbol{F}}_{T}\big]))\big),
    \end{cases}
\end{equation}
where $\boldsymbol{F_i}$ denotes the feature map of the $i$-th frame within a temporal window $T=5$, and $\mathcal{W}(\cdot)$ denotes the wavelet-based frequency enhancement operation that combines multi-level decomposition, depth-wise filtering, and inverse wavelet reconstruction.

\subsection{Residual Compensation Unit}
Our learning framework jointly exploits spatial, temporal, and frequency-domain cues for belt-crack representation. In practice, however, discrepancies may arise because these domains are learned through different pathways. To address this issue, we design the RCU to dynamically align and enhance complementary characteristics across the three domains.

As illustrated in Fig. \ref{fig:arch} and formalized in Eq. (\ref{rcu_triple}), the RCU is implemented using a stack of \textit{channel-spatial attention blocks} (CSABs). Each CSAB applies channel attention (CAB) followed by spatial attention (SAB) to a convolutionally transformed input and adds a residual connection to stabilize optimization. Specifically, for the local temporal-frequency feature $\boldsymbol{F_{f_{l}t}}$ and the spatial-temporal feature $\boldsymbol{F_{st}}$, we first concatenate them along the channel dimension and then feed the result into $n$ cascaded CSABs:
\begin{equation}
    \label{eq9}
    \begin{cases}
        \boldsymbol{\widehat{F_{e}}}  = Concat\left ( \left [\boldsymbol{ F_{f_{l}t}} ,\boldsymbol{F_{st}}  \right ]  \right ), \\
        CSAB\left ( \boldsymbol{X} \right ) = SAB\left ( CAB\left ( Conv\left ( \boldsymbol{X}  \right )   \right )  \right ) +\boldsymbol{X},  \\
        \boldsymbol{F_{f_{l}st}}  = CSAB_{n} \left ( CSAB_{n-1}...\left ( CSAB_{1}\left ( \boldsymbol{\widehat{F_{e}}} \right )  \right )  \right )  ,
    \end{cases}
\end{equation}
where $n$ is the number of CSABs. $\boldsymbol{F_{ft}}$ is processed in the same operations as $\boldsymbol{F_{f_{l}t}}$ and $\boldsymbol{F_{f_{g}t}}$. The final crack representation $\boldsymbol{F_{fst}}$ is obtained by combining the compensated outputs from the local and global branches (as Eq. (\ref{rcu_triple})). To construct $\boldsymbol{F_{f_{l}t}}$, we use local frequency features $\boldsymbol{F_{F}}$ with temporal features $\boldsymbol{F_{T}}$ using a lightweight convolutional fusion. For $\boldsymbol{F_{f_{g}t}}$, a \textit{Swin Transformer} (Swin) block is applied to capture global temporal–frequency context. This design highlights salient regions, retains critical details, and reduces domain gaps.

\subsection{Loss Function}
To address severe class imbalance and large object-scale variations in our datasets, we adopt the composite loss from YOLOX\citep{Ge2021YOLOXEY} and further optimize the regression term. The overall objective loss is:
\begin{equation}
    \label{loss-total}
        \mathcal{L}  = \lambda_{reg}{\mathcal{L}_{reg}}  + \lambda_{cls} {\mathcal{L}_{cls}} + \lambda_{obj} {\mathcal{L}_{obj}}   
\end{equation}
where $\mathcal{L}_{reg}, \mathcal{L}_{cls}$, and $\mathcal{L}_{obj}$ denote the box regression, classification, and objectness losses, respectively. $\lambda_{reg}$, $\lambda_{cls}$, and $\lambda_{obj}$ are scalar weights balancing the three terms.

For $\mathcal{L}_{obj}$ and $\mathcal{L}_{cls}$, we adopt sigmoid focal loss \citep{Lin2017} to alleviate class imbalance between positive and negative samples. Specifically, the total loss in Eq. (\ref{loss-total}) is further formulated as:
\begin{equation}
    \label{loss-detail}
    \begin{cases} 
    \mathcal{L}_{reg} = \zeta   \mathcal{L}_{iou} + \eta \mathcal{L}_{nwd}  , \\
    NWD\left ( {N}_{p} ,{N}_{g}\right )  = exp\left ( -\frac{\sqrt{W_{2}^{2}\left ( {N}_{p} ,{N}_{g} \right )  } }{C}  \right ) ,
    \end{cases}
\end{equation}
where $\mathcal{L}_{iou}=1-IoU\left ( {B}_{p},{B}_{g} \right )$ and $\mathcal{L}_{nwd} =  1- NWD\left ( {N}_{p} ,{N}_{g}\right )$. The \textit{Normalized Wasserstein Distance} (NWD) loss is incorporated into $\mathcal{L}_{reg}$ to better model the spatial distribution characteristics of belt cracks. Here, ${C}$ is a dataset-related constant. $ B_p$, $B_g$ denote the predicted and ground-truth box, respectively, while $N_p$ and $N_g$ are their corresponding Gaussian distribution.

\section{Experiments}
\label{sec5}
\subsection{Datasets and Evaluation Metrics}
To rigorously validate the effectiveness of our datasets and evaluate the superiority of BeltCrackDet, we conduct comprehensive experiments on both BeltCrack14ks and BeltCrack9kd, with the four classic metrics: Precision, Recall, F1, and \textit{mean Average Precision} (mAP$_{50}$).


\subsection{Implementation Details}
Specifically, the sliding window size of sequence $T$ is set to 5, 
and all methods adopt a fixed input resolution of $512\times 512$ for fair comparison. BeltCrackDet is trained for 100 epochs with a batch size of 4 using the \textit{Stochastic Gradient Descent} (SGD) optimizer (initial learning rate 0.01, momentum 0.937) and a cosine annealing schedule for stable convergence. 

The balancing factors $\lambda_{reg}$, $\lambda_{cls}$, and $\lambda_{obj}$ in Eq. (\ref{loss-detail}), are set to 5, 1, and 1, respectively, with $\zeta =\eta =0.5$. During inference, \textit{Non-Maximum Suppression} (NMS) is applied with an 
\textit{Intersection-over-Union} (IoU) threshold of 0.65 and a confidence threshold of 0.001. All experiments are implemented in Pytorch on two NVIDIA GeForce RTX 4090 GPUs. Each experiment is repeated three times to ensure reliability.

\subsection{Comparisons with Other Methods}
To comprehensively evaluate our datasets and the BeltCrackDet baseline, we compare our method with 12 existing detection methods, covering both single-image and sequential-frame categories, as summarized in Table \ref{table_qua_com}.

\begin{table*}[ht]
\centering
\renewcommand{\arraystretch}{1.0} 
\caption{Quantitative comparisons on two datasets. The top-three results are highlighted by \textbf{\underline{underlined bold}}, \textbf{bold} and \underline{underlined}, respectively. AW=Adaptive Weighting, and RCU=Residual Compensation Unit.}
\label{table_qua_com}
\resizebox{\linewidth}{!}{
\begin{tabular}{c|cr|cccc|cccc}
\hline
\multirow{2}{*}{\textbf{Category}} & 
\multirow{2}{*}{\textbf{Methods}} & 
\multirow{2}{*}{\textbf{Publication}} & 
\multicolumn{4}{c}{\textbf{BeltCrack14ks}}  & 
\multicolumn{4}{c}{\textbf{BeltCrack9kd}}    \\ \cline{4-11} 
&  & & $\textbf{mAP}_{\textbf{50}}$  & \textbf{Precision} & \textbf{Recall} & \textbf{F1}     
& $\textbf{mAP}_{\textbf{50}}$  & \textbf{Precision} & \textbf{Recall} & \textbf{F1}     
\\ \hline
\multicolumn{1}{c|}{\multirow{9}{*}{Single-image}} & RIND\citep{RINDNet}                 &  ICCV'21          & 52.25 &  83.57    & 63.72 & 72.36 & 24.34 & 48.62    & 44.48 & 46.91 \\
\multicolumn{1}{c|}{}                              & YOLOX\citep{Ge2021YOLOXEY}                   &  arXiv'21         & 61.79 & 81.01    &  \underline{74.65} & 79.29 & 24.82 & \underline{50.54}   & 49.36 & 50.22 \\
\multicolumn{1}{c|}{}                              & CrackFormer\citep{CrackFormer}             &  ICCV'21         & 48.71 & 75.57    & 65.56 & 70.21 & 4.17 & 12.76    & 33.88 & 18.54 \\
\multicolumn{1}{c|}{}                              & JTFN\citep{JTFN}                    &  ICCV'21          & 40.15 & 65.23    & 61.79 & 64.02 & 0.66 & 1.38    & 32.03 & 2.65 \\
\multicolumn{1}{c|}{}                              & PyramidFlow\citep{PyramidFlow}             &  CVPR'23          & 60.85 & 81.63    &  \textbf{\underline{{75.66}}} & 78.85 & 23.86 & 49.67    &  \underline{53.72} &  \textbf{52.81} \\
\multicolumn{1}{c|}{}                              & CrackNex\citep{CrackNex}                &  ICRA'24         &  62.59 & 82.51    & 74.47 &  79.82 & 24.79 & 50.23    & 52.95 & 50.75 \\
\multicolumn{1}{c|}{}                              & SimCrack\citep{SimCrack}                &  WACV'24          &  \underline{64.33} &  87.39   & 73.28 &  \underline{80.84} &  24.87 &  \textbf{50.79}   & 52.18 & 48.67 \\
\multicolumn{1}{c|}{}                              & ADL4VAD\citep{ADL4VAD}                 &  WACV'25        & 52.04 & 79.51    & 61.69 & 71.84 & 21.76 & 47.86    & 47.06 & 45.37 \\
\multicolumn{1}{c|}{}                              & UniAS\citep{UniAS}                   &  WACV'25           & 51.63 & 81.47    & 57.58 & 70.56 & 21.36 & 48.07    & 40.25 & 52.35 
\\ \hline

\multicolumn{1}{c|}{\multirow{4}{*}{Sequential}}                  &vid-TLDR\citep{choi2024vid} & CVPR'24  & 59.11 & 80.89 & 67.72 & 75.94 & 22.42 & 49.35 & 48.12 & 48.18 \\

& TAPTR\citep{li2024taptr} &ECCV'24 &63.38 & 84.52 & 74.11 & 79.09 & \underline{25.06} & 48.35 & 53.14 & 51.31
\\

& AOST\citep{yangAOST2024} & TPAMI'24 & \textbf{65.22} & \textbf{89.71} & 73.96 & \textbf{81.45} & \textbf{26.33} & 47.17 & \textbf{54.38} & \underline{52.54} 
\\

& \textbf{BeltCrackDet} (w/ AW)  & \textbf{————}   &64.26 &\underline{88.83} &73.37 &80.33 &24.73 &47.51 &51.26 &49.72   \\

& \textbf{BeltCrackDet} (w/ RCU)           & \textbf{————}                  & \textbf{\underline{67.08}} &  \textbf{\underline{{91.03}}}    &  \textbf{74.89} &  \textbf{\underline{{82.17}}} &  \textbf{\underline{30.79}} &  \textbf{\underline{61.69}}    &  \textbf{\underline{54.94}} &  \textbf{\underline{56.98}} 

\\ \hline
\end{tabular}}

\begin{tablenotes}
\scriptsize
\item \textbf{Note:} Results report the best-performing outcomes. Three independent runs were performed to evaluate stability; our BeltCrackDet (w/ RCU) achieves $81.88 \pm 0.40$ (F1) on BeltCrack14ks and $56.92 \pm 0.07$ (F1) on BeltCrack9kd. Similarly, the strongest baseline (AOST) obtains $80.27 \pm 1.17$ (F1) and $51.95 \pm 0.84$ (F1), respectively.
\end{tablenotes}

\end{table*}

(1) {Quantitative Comparisons} 

Table \ref{table_qua_com} presents quantitative comparisons of 13 detection methods (including ours) on both BeltCrack14ks and BeltCrack9kd. Based on these results, three main observations can be easily drawn.

The first finding is that the proposed datasets are \textit{usable and practically effective} for industrial belt crack detection. All compared models can be successfully trained and evaluated on our datasets without any compatibility issues, indicating their good robustness and generalizability. The results in Table \ref{table_qua_com} also demonstrate that our datasets provide sufficient discriminative capacity to reveal the performance gaps among different models.
For example, on BeltCrack14ks, JTFN \citep{JTFN} achieves an mAP$_{50}$ 40.15\% and an F1 64.02\%, while our BeltCrackDet attains the highest mAP$_{50}$ 67.08\% and F1 82.17\%. Similarly, on BeltCrack9kd, JTFN \citep{JTFN} yields only 0.66\% mAP$_{50}$ and 2.65\% F1, confirming the effectiveness of our datasets in challenging real-world scenarios.

The second finding is that our datasets ($i.e.$, BeltCrack14ks and BeltCrack9kd) are challenging for all methods, with BeltCrack9kd being the more difficult one.
On BeltCrack14ks, the best-performing method across all methods achieves only mAP$_{50}$ 67.08\%, Precision 91.03\%, Recall 74.89\% and F1 82.17\%. The difficulty is even more apparent on BeltCrack9kd, where the SOTA metrics, $i.e.$, the mAP$_{50}$ 30.79\%, Precision 61.69\%, Recall 54.94\% and F1 56.98\% by our method, are markedly lower than those on BeltCrack14ks. One possible reason is the shorter average sequence length in BeltCrack9kd (see the Img.$_{avg/seq}$ in Table \ref{table_ourdataset}), which provides less temporal context for learning. In addition, faster belt motion increases motion blur and occlusion, further increasing task difficulty.

The third observation is that our baseline achieve SOTA performance on both datasets. On BeltCrack14ks, it attains an mAP$_{50}$ 67.08\%, exceeding the second-best method AOST \citep{yangAOST2024} $65.22\%$ by $1.86\%$. It also achieves a peak precision $91.03\%$, while the runner-up reaches $89.71\%$. On the more challenging BeltCrack9kd, although overall performance decreases for all methods, BeltCrackDet still achieves the best mAP$_{50}$ (30.79\%) and Precision (61.69\%), surpassing AOST \citep{yangAOST2024}. Moreover, our baseline also yields the top Recall (54.94\%) and F1 (56.98\%) among all compared approaches.

\begin{figure*}
	\centering
    \includegraphics[width=0.76\textwidth]{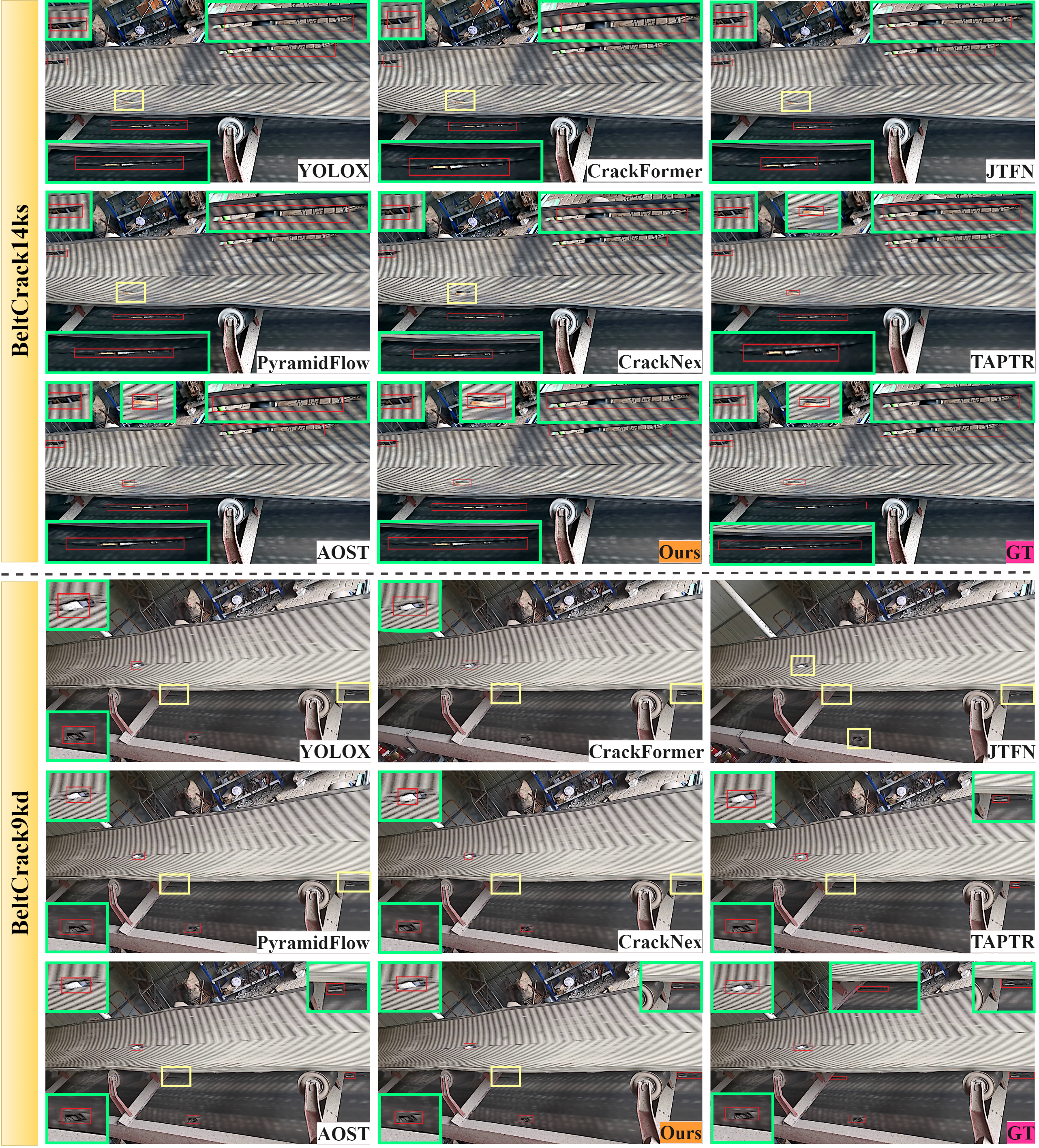}
	\caption{Visual comparison. GT=Ground Truth. Red boxes represent detected cracks, light-yellows ones denote missed cracks, and the cyan boxes highlight the zoomed-in areas.}
	\label{fig:vis_SD}
\end{figure*}

The last finding is that sequential methods consistently surpass single-image ones, particularly on the challenging BeltCrack9kd (Table \ref{table_qua_com}), where our BeltCrackDet leads with 30.79\% mAP$_{50}$, and 56.98\% F1. Notably, representative single-image methods nearly fail, e.g., CrackFormer \citep{CrackFormer} (mAP$_{50}$ 4.17\%) and JTFN \citep{JTFN} (mAP$_{50}$ 0.66\%). These failures mainly stem from belt-specific imaging factors (e.g., motion blur, deformation, and complex textures) rather than model capacity alone. Fast belt motion causes blur and occlusion that obscure the sharp edge cues required by static detectors. As a result, CrackFormer \citep{CrackFormer}, whose attention modeling relies on clear structural cues, becomes unreliable when boundaries are blurred. Periodic belt deformation further distorts crack geometry, making topology-refinement methods such as JTFN \citep{JTFN} less stable when structural evidence is inconsistent. In addition, low-saliency cracks embedded in repetitive textures are difficult to distinguish without temporal context. In contrast, our triple-domain design integrates temporal continuity, frequency-aware representation, and spatial detail modeling, enabling complementary evidence aggregation beyond single-frame structural cues.

Furthermore, we conducted supplementary experiments to analyze the detection accuracy of BeltCrackDet under these challenging conditions. Specifically, we evaluated an adaptive-weighting (AW) variant to enhance discriminative feature learning. However, both quantitative and visual analysis reveal that its results seem far lower than those obtained with RCU (in Table \ref{table_qua_com}). These comparisons indicate that the RCU in our BeltCrackDet is more effective than standard adaptive weighting for feature fusion.

(2) {Visualization Comparisons}

To illustrate the detection performance of different methods on the two datasets, we conduct qualitative examples in Fig. \ref{fig:vis_SD}. Two key observations could be found.

First, both datasets enable belt-crack detection, further supporting their practical validity. All compared methods can detect cracks in the presented samples, although their detection quality varies. For instance, some methods, such as BeltCrackDet, AOST and TAPTR, can detect nearly all cracks in a sample from BeltCrack14ks, whereas others, $e.g.$,  CrackFormer and JTFN, miss some real cracks. Such visual results reveal performance differences among models for industrial belt crack detection.

Second, detecting all belt cracks remains challenging, especially on the more complex dataset, BeltCrack9kd. For example, even on BeltCrack14ks, the models including TAPTR, AOST, and our BeltCrackDet, can almost completely detect all cracks, while the others such as YOLOX often tend to miss true cracks ($i.e.$, missed detections). On BeltCrack9kd, the difficulty further increases, and most methods struggle to correctly localize subtle or faint cracks in samples. One possible explanation is that severe occlusion, weak illumination, and small crack sizes reduce the availability of discriminative features, thereby limiting the detection capability of existing methods.

(3) {Precision-Recall Curves}

\begin{figure}
	\centering
	\includegraphics[width=1.0\columnwidth]{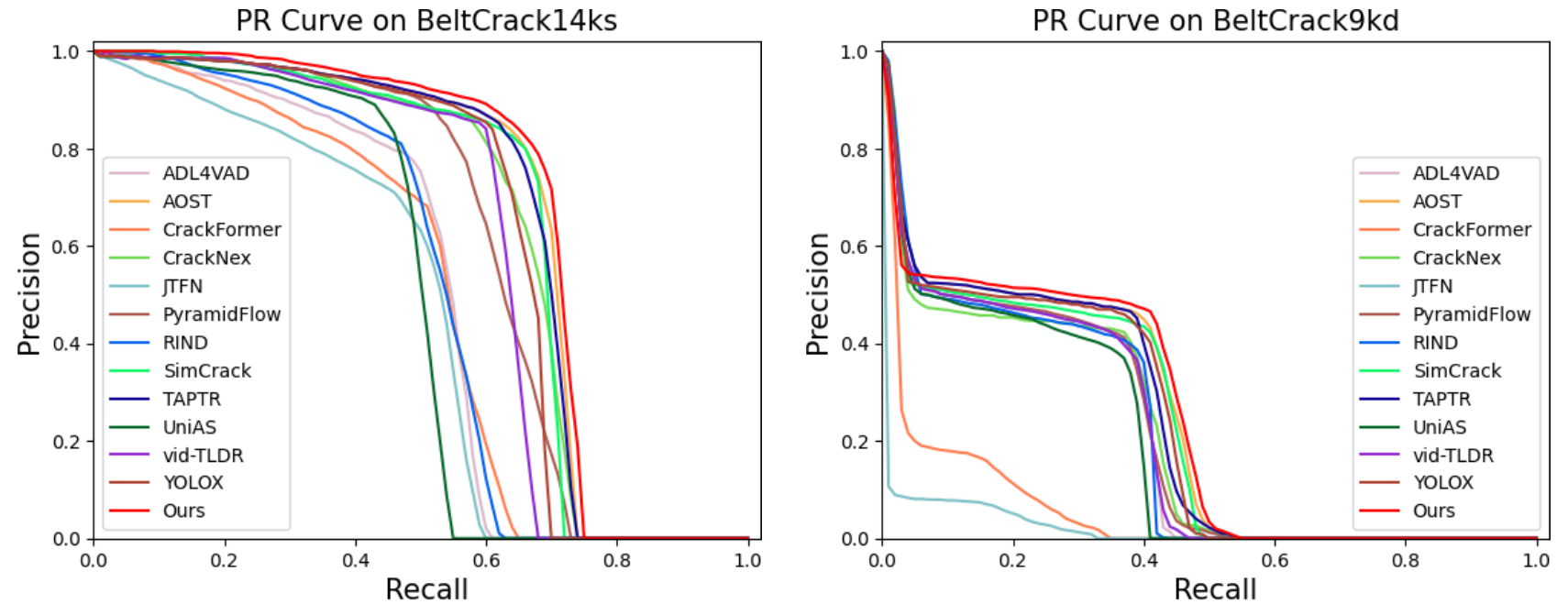}
	\caption{The PR curves of 12 representative methods on datasets BeltCrack14ks and BeltCrack9kd.}
	\label{fig:pr_SD}
\end{figure}

To further evaluate the comprehensive detection performance of our baseline, we conduct experiments to generate the \textit{Precision-Recall} (PR) curves for various methods on both datasets, as shown in Fig. \ref{fig:pr_SD}. 

The  first observation is that all compared detection methods can be trained and evaluated on both datasets, although their detection performance varies. The PR curves show clear differences in the \textit{Area Under the Curve} (AUC) across methods. Notably, the AUC of the same model on BeltCrack9kd is significantly smaller than that on BeltCrack14ks, indicating weaker overall performance. This further suggests that BeltCrack9kd is more challenging, due to its higher visual and contextual complexity.

The second observation is that our proposed BeltCrackDet consistently outperforms the others on both datasets. Its PR curve is concentrated toward the top-right region, corresponding to the highest AUC and stronger overall detection capability. These results further demonstrate the robustness and superiority of our baseline for industrial belt cracks detection.

(4) {Complexity Comparisons}

Beyond detection performance, we further compare the model complexity of BeltCrackDet and 12 representative methods. Since model complexity is dataset-independent, experiments are conducted only on BeltCrack14k. Quantitative comparisons of FLOPs, Params and FPS are summarized in Table \ref{complexity}.

\begin{table}
\centering
\renewcommand{\arraystretch}{1.5}
\caption{Complexity comparisons on BeltCrack14ks, with the best result marked in \textbf{bold}.}
\scriptsize                      
\resizebox{\linewidth}{!}{
\begin{tabular}{l|ccc|rrr}
\hline
\textbf{Methods}       & \textbf{Frames}   & $\textbf{mAP}_\textbf{50}$ $\uparrow$ & \textbf{F1}  $\uparrow $   & \textbf{FLOPs}  $\downarrow $  & \textbf{Params}  $\downarrow $& \textbf{FPS} $ \uparrow $  \\ \hline
RIND\cite{RINDNet}       & single   & 52.25 & 72.36 & 265.15G  & \textbf{8.37M}  & 7.6  \\
YOLOX\cite{Ge2021YOLOXEY}         & single   & 61.79 & 79.29 & \textbf{29.68G}   & 9.2M   & \textbf{50.8} \\
CrackFormer\cite{CrackFormer}   & single   & 48.71 & 70.21 & 50.58G   & 8.42M  & 17.3  \\
JTFN\cite{JTFN}          & single   & 40.15 & 64.02 & 656.48G  & 17.87M & 1.8  \\
PyramidFlow\cite{PyramidFlow}   & single   & 60.85 & 78.85 & 76.28G   & 10.27M & 14.5 \\
CrackNex\cite{CrackNex}      & single   & 62.59 & 79.82 & 47.35G   & 14.21M & 13.6  \\
SimCrack\cite{SimCrack}      & single   & 64.33 & 80.84 & 76.19G   & 9.54M  & 15.1  \\
ADL4VAD\cite{ADL4VAD}       & single   & 52.04 & 71.84 & 120.72G  & 15.88M & 11.7  \\
UniAS\cite{UniAS}         & single   & 51.63 & 70.56 & 127.02G  & 11.95M & 12.9  \\
vid-TLDR\cite{choi2024vid}      & multiple & 59.11 & 75.94 & 351.14G  & 46.71M & 2.6  \\
TAPTR\cite{li2024taptr}         & multiple & 63.38 & 79.09 & 57.28G   & 27.40M & 13.3 \\
AOST\cite{yangAOST2024}         & multiple & 65.22 & 81.45 & 182.45G  & 33.84M & 9.5  \\
\textbf{BeltCrackDet} (ours) & \textbf{multiple} & \textbf{67.08} & \textbf{82.17} & 70.42G & 20.70M & 16.7  \\ \hline
\end{tabular}
\label{complexity}
}
\end{table}

First, although our BeltCrackDet is a sequential model, its FLOPs and Params remain moderate among competitors. 
Specifically, it requires 70.42G FLOPs with 20.70M parameters.
By contrast, JTFN \citep{JTFN} incurs a markedly higher cost (656.48G FLOPs), while YOLOX \citep{Ge2021YOLOXEY} has fewer FLOPs (29.68G). In terms of parameters, our model’s 20.70M lies between vid-TLDR \citep{choi2024vid} (46.71M) and RIND \citep{RINDNet}, which is the smallest (only 8.37M). A possible explanation for the higher cost compared to single-image baselines is the additional frequency-domain modeling and fusion modules.

Second, as computational cost rises, the FPS of BeltCrackDet decreases to 16.7, which is lower than that of most compared methods. For example, YOLOX has low complexity (29.68G FLOPs, 9.2M parameters) and high FPS (50.8), but its accuracy on BeltCrack14ks is limited (mAP$_{50}$ 61.79\%, F1 79.29\%). On the same dataset, our baseline achieves substantially better detection quality (mAP$_{50}$ 67.08\%, F1 82.17\%). In real-world applications, sacrificing some FPS in exchange for improved detection accuracy is often worthwhile \citep{YOLO-MS}.

(5) {Crack Heatmap Comparisons}

\begin{figure}
	\centering
	\includegraphics[width=1.0\linewidth]{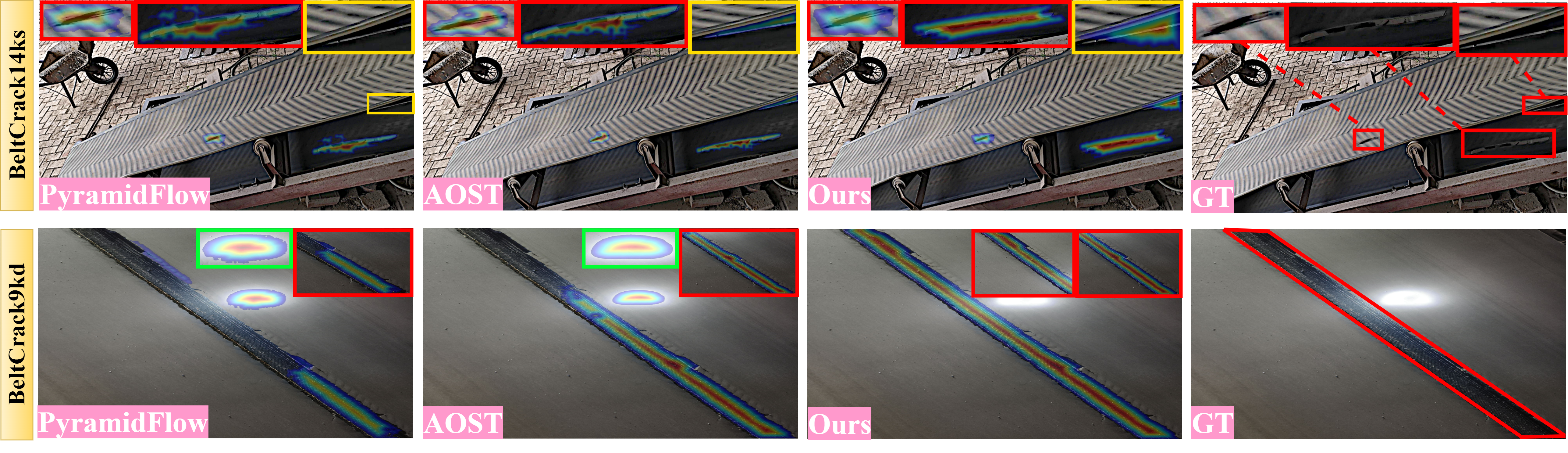}
	\caption{Heatmap visualization comparisons. Yellow boxes indicate the missed cracks, red boxes denote the crack heatmaps and their zoomed presentations, while cyan boxes highlight false detections.}
	\label{fig:heatmap}
\end{figure}

To further show interpretability, we compute belt-crack heatmaps for three representative methods and present two examples (one sample per dataset) in Fig. \ref{fig:heatmap}.

From these visualizations, several consistent observations emerge. First, on both datasets, the heatmaps produced by BeltCrackDet better capture crack location and shapes, with fewer missed detections. For example, on BeltCrack14ks, PyramidFlow \citep{PyramidFlow} fails to highlight a subtle crack ($i.e.$, a miss), whereas our baseline correctly activates the corresponding region. Second, even on the more challenging BeltCrack9kd, our method yields higher-quality and more reliable heatmap than PyramidFlow \citep{PyramidFlow} and AOST \citep{yangAOST2024}. Both PyramidFlow and AOST exhibit false detections (marked by cyan boxes), where non-crack regions are incorrectly identified. In contrast, our heatmap seems more consistent (including location and shape) with real cracks, effectively suppressing false responses.
These qualitative findings are exactly consistent with the quantitative comparisons in Table \ref{table_qua_com}.

\subsection{Ablation Study}

(1) {Effect of Different Modules}

We first study how different module configurations affect crack detection performance. In Table \ref{module_ablation}, two consistent observations can easily be drawn.

\begin{table}[ht]
\centering
\renewcommand{\arraystretch}{1.8}
\caption{The ablation on different module assembling. Specifically, R(A) only applies RCU$_1$, R(B) contains both RCU$_1$ and RCU$_2$, and R(C) employs all RCU components. The best metrics are marked in \textbf{bold}.
}
\label{module_ablation}
\resizebox{\linewidth}{!}{
\begin{tabular}{llcccccccccccccc}
\hline
\multicolumn{2}{l|}{}       &   &    &    &  &         & \multicolumn{1}{c|}{}   & \multicolumn{4}{c}{\textbf{BeltCrack14ks}}                                        & \multicolumn{4}{c}{\textbf{BeltCrack9kd}}          
\\ \cline{9-16} 
\multicolumn{2}{l|}{\multirow{-2}{*}{\textbf{Settings}}} & \multirow{-2}{*}{\textbf{HSM}}    & \multirow{-2}{*}{\textbf{ATM}}    & \multirow{-2}{*}{\textbf{WFM}}    & \multirow{-2}{*}{\textbf{R(A)}}   & \multirow{-2}{*}{\textbf{R(B)}}   & \multicolumn{1}{c|}{\multirow{-2}{*}{\textbf{R(C)}}}   &  $\textbf{mAP}_{\textbf{50}}$                        & \textbf{Precision}            & \textbf{Recall}               & \multicolumn{1}{c|}{\textbf{F1}}    &  $\textbf{mAP}_{\textbf{50}}$                & \textbf{Precision}            & \textbf{Recall}               & \textbf{F1}                   
\\ \hline
\multicolumn{2}{l|}{w/o All}                    & {-} & { -} & { -} & { -} & { -} & \multicolumn{1}{c|}{{ -}} & 61.01                        & 87.53                & 70.76                & \multicolumn{1}{c|}{78.26} & 20.06                & 46.35                & 41.73                & 44.18                
\\
\multicolumn{2}{l|}{w HSM}                      & {\ding{52}} & { -} & { -} & { -} & { -} & \multicolumn{1}{c|}{{ -}} & 63.73                        & 88.14                 & 73.27                & \multicolumn{1}{c|}{80.01} & 20.48                & 46.97                & 43.55                & 45.29                
\\
\multicolumn{2}{l|}{w HSM \&ATM}                & {\ding{52}} & {\ding{52}} & { -} & { -} & { -} & \multicolumn{1}{c|}{{ -}} & {64.02} & 89.65                & 72.62                & \multicolumn{1}{c|}{80.24} & 22.19                & 45.73                & 49.19                & 47.42               
\\
\multicolumn{2}{l|}{w/o R}                      & {\ding{52}} & {\ding{52}} & {\ding{52}} & { -} & { -} & \multicolumn{1}{c|}{{ -}} & {65.16} & 89.89                & 73.92                & \multicolumn{1}{c|}{81.13} & 23.36                & 47.18                & 50.08                & 48.59                \\
\multicolumn{2}{l|}{w R(A)}                     & {\ding{52}} & {\ding{52}} & {\ding{52}} & {\ding{52}} & { -} & \multicolumn{1}{c|}{{ -}} & 66.37                        & 90.09                & 74.22                & \multicolumn{1}{c|}{79.72} & 24.58                & 48.62                & 51.77                & 50.14                
\\
\multicolumn{2}{l|}{w R(B)}                     & {\ding{52}} & {\ding{52}} & {\ding{52}} & { -} & {\ding{52}} & \multicolumn{1}{c|}{{ -}} & 66.76                        & 90.31                 & \textbf{75.13}               & \multicolumn{1}{c|}{80.12} & 25.78                & 49.95                 & 53.13                & 51.57                
\\
\multicolumn{2}{l|}{w R(C)}                     & {\ding{52}} & {\ding{52}} & {\ding{52}} & { -} & { -} & \multicolumn{1}{c|}{{\ding{52}}} & \textbf{67.08}                        & \textbf{91.03}                & 74.89                & \multicolumn{1}{c|}{\textbf{82.17}} & \textbf{30.79}                & \textbf{61.69}                & \textbf{54.94}                & \textbf{56.98}                
\\ \hline
\end{tabular}
}
\end{table}

First, each individual module contributes positively. On BeltCrack14ks, the baseline without any module (w/o All) yields mAP$_{50}$ 61.01\% and F1 78.26\%. Adding HSM improves performance to mAP$_{50}$ 63.73\% and  F1 80.01\%, while further incorporating ATM on top of HSM further lifts mAP$_{50}$ to 64.02\% and F1 to 80.24\%. Moreover, utilizing the frequency-domain branch without RCU could also bring clear gains, reaching to mAP$_{50}$ 65.16\% and F1 81.13\%.

Second, the full configuration consistently secures the best results. For example, on BeltCrack14ks, only the configuration ``w R(C)'' could obtain the best mAP$_{50}$ (67.08\%) and F1 (82.17\%).
On BeltCrack9kd, the same scheme improves these two metrics to 30.79\% and 56.98\%, respectively. These findings indicate that the proposed modules are complementary. Only when all modules are combined together, our method could achieve the best detection accuracy.

(2) {Effect of Different Regression Losses}

We evaluate the impact of the regression loss ${\mathcal{L}_{reg}}$ by comparing three variants (Table \ref{loss_ablation}). Here, ``${\mathcal{L}_{nwd}}$'' denotes setting the loss in Eq. (\ref{loss-detail}) to ``${\mathcal{L}_{nwd}}$'' without ${\mathcal{L}_{iou}}$. ``${\mathcal{L}_{iou}}$'' represents that ${\mathcal{L}_{reg}}$ only uses ${\mathcal{L}_{iou}}$. Moreover, ``${\mathcal{L}_{reg}}$'' combines ${\mathcal{L}_{iou}}$ and ${\mathcal{L}_{nwd}}$ to jointly balance localization accuracy and geometric consistency. 

\begin{table}[ht]
\centering
\renewcommand{\arraystretch}{1.5} 
\caption{Three loss function comparisons, with the best metrics marked in \textbf{bold}.}
\label{loss_ablation}
\resizebox{1.0\columnwidth}{!}{
\begin{tabular}{l|cccccccc}
\hline
\multicolumn{1}{l|}{\multirow{2}{*}{\textbf{Settings}}} & \multicolumn{4}{c}{\textbf{BeltCrack14ks}}                                           & \multicolumn{4}{c}{\textbf{BeltCrack9kd}}                                           \\ \cline{2-9} 
\multicolumn{1}{l|}{}                          & $\textbf{mAP}_\textbf{50}$                     & \textbf{Precision} & \textbf{Recall} & \multicolumn{1}{c|}{\textbf{F1}}    & $\textbf{mAP}_\textbf{50}$                      & \textbf{Precision} & \textbf{Recall} & \textbf{F1}    \\ \hline
${\mathcal{L}_{nwd}}$                                           & 65.77 & 87.42     & \textbf{78.26}  & \multicolumn{1}{l|}{80.14} & 23.72 & 47.65     & 53.41  & 50.08                     \\
${\mathcal{L}_{iou}}$                                           & 66.39 & \textbf{92.11}     & 72.19  & \multicolumn{1}{l|}{80.35} & 24.49 & 50.68     & 53.15  & 50.73                     \\
${\mathcal{L}_{reg}}$ (with ${\mathcal{L}_{nwd}}$ and ${\mathcal{L}_{iou}}$) & \textbf{67.08}                     & 91.03     & 74.89  & \multicolumn{1}{c|}{\textbf{82.17}} & \textbf{30.79}                     & \textbf{61.69}     & \textbf{54.94}  & \textbf{56.98} \\ \hline
\end{tabular}
}
\end{table}

From Table \ref{loss_ablation}, two conclusions could be drawn. One is that the optimized ${\mathcal{L}_{reg}}$ consistently enables our BeltCrackDet to achieve the most comprehensive performance. For instance, ${\mathcal{L}_{reg}}$ achieves a peak mAP$_{50}$ of 67.08\% and an F1 of 82.17\% on BeltCrack14ks. Meanwhile, it also attains the highest metrics (including mAP$_{50}$ 30.79\%, Precision 61.69\%, Recall 54.95\% and F1 56.98\%) on BeltCrack9kd.

The other is that both ${\mathcal{L}_{nwd}}$ and ${\mathcal{L}_{iou}}$ contribute positively during training. Specifically, when trained by the former, BeltCrackDet reaches an mAP$_{50}$ 65.77\% and F1 80.14\% on BeltCrack14ks, only 1.31\% and 2.03\% lower than the best mAP$_{50}$ and F1 achieved by ${\mathcal{L}_{reg}}$, respectively.
These comparisons quantitatively imply that ${\mathcal{L}_{reg}}$ is the most effective loss design for our BeltCrackDet.

(3) {Effect of Different Hyperparameters}

\begin{figure}
	\centering
	\includegraphics[width=1.0\columnwidth]{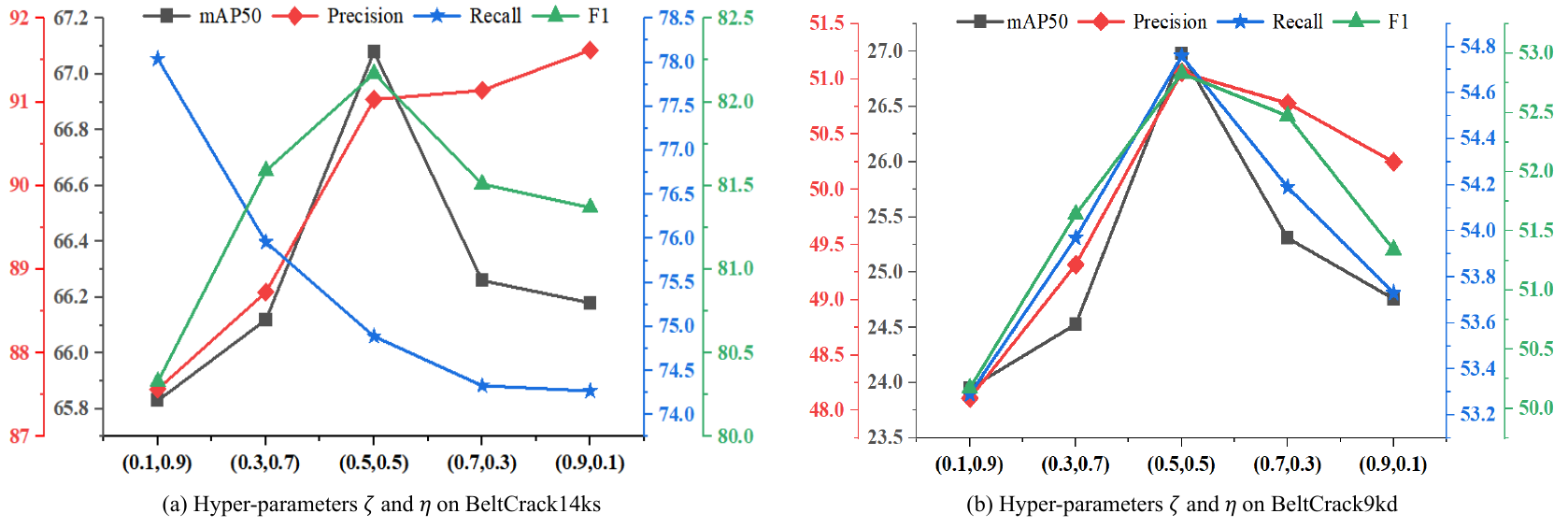}
	\caption{The effects of hyperparameters $\zeta$ and $\eta$ on our baseline method.}
	\label{fig:hyper-para}
\end{figure}

To evaluate the effect of the two primary hyperparameters $\zeta$ and $\eta$ in Eq. (\ref{loss-detail}), we conduct two groups of comparative experiments, with the corresponding results illustrated in Fig. \ref{fig:hyper-para}.

In these experiments, $\zeta$ is incrementally increased with a step size of 0.2, while $\eta$ is decreased by the same step size (0.2), and all other settings kept constant. These results clearly indicate that both parameters significant impact the overall detection performance. Notably, the best performance on both datasets is achieved only when $\zeta$ = 0.5 and $\eta$ = 0.5, suggesting a balanced contribution between these two terms. In contrast, for our method, any other hyperparameter combinations ($\zeta, \eta$) will always cause an obvious performance drop.

We also conduct an ablation study on the loss-term coefficient $(\lambda_{reg}, \lambda_{cls}, \lambda_{obj})$, defined in Eq. (\ref{loss-total}), to examine their impact on performance. Specifically, these coefficients are mutable, while all other settings are frozen, as shown in Table \ref{loss_parameter}. 

\begin{table}
    \centering
    \renewcommand{\arraystretch}{1.1} 
    \caption{Ablation of $(\lambda_{{reg}},\lambda_{{cls}},\lambda_{{obj}})$ on BeltCrack14ks, with the best metrics marked in \textbf{bold}.}
    \label{loss_parameter}
    \resizebox{1.0\columnwidth}{!}{
    \begin{tabular}{c|ccccccccc}
    \hline
    \textbf{Settings}         & \multicolumn{9}{c}{\textbf{BeltCrack14ks}}                            \\ \cline{2-10} 
{${(\lambda_{reg}, \lambda_{cls}, \lambda_{obj})}$} & \textbf{mAP$_{50}$}          & \textbf{Precision}      & \textbf{Recall}         & \multicolumn{1}{c|}{\textbf{F1}}             & \multicolumn{1}{c|}{{${(\lambda_{reg}, \lambda_{cls}, \lambda_{obj})}$}} & \textbf{mAP$_{50}$} & \textbf{Precision} & \textbf{Recall} & \textbf{F1}    \\ \hline
(1, 1, 1)          & 63.17          & 87.33          & 71.16          & \multicolumn{1}{c|}{78.25}          & \multicolumn{1}{c|}{(5, 0.5, 1)}        & 66.12 & 89.25     & 74.01  & 78.87 \\
(3, 1, 1)          & 65.14          & 88.85          & 73.74          & \multicolumn{1}{c|}{81.06}          & \multicolumn{1}{c|}{(5, 2, 1)}          & 66.74 & 90.37     & 75.12  & 80.06 \\
\textbf{(5, 1, 1)} & \textbf{67.08} & \textbf{91.03} & \textbf{74.89} & \multicolumn{1}{c|}{\textbf{82.17}} & \multicolumn{1}{c|}{(5, 1, 0.5)}        & 63.76 & 88.31     & 73.26  & 78.48 \\
(7, 1, 1)          & 66.24          & 89.96          & 74.17          & \multicolumn{1}{c|}{79.12}          & \multicolumn{1}{c|}{(5, 1, 2)}          & 66.49 & 90.11     & 74.21  & 80.82 \\ \hline
\end{tabular}
}
\end{table}

Notably, when the coefficients are set to (5, 1, 1), BeltCrackDet achieves the highest mAP$_{50}$ (67.08\%) and F1 (82.17\%) on BeltCrack14ks. This configuration emphasizes the regression term, helping the model capture more precise bounding-box localization while maintaining classification accuracy. In contrast, increasing or decreasing $\mathbf{\lambda_{reg}}$ on this point could cause noticeable declines in both mAP$_{50}$ and F1. These results indicate that (5, 1, 1) could often provide an effective balance between localization precision and detection robustness.

(4) {Effect of Wavelet Decomposition Levels}

To evaluate the impact of wavelet decomposition levels on crack detection performance, experiments are also conducted across five settings on both datasets.

\begin{table}[ht]
\caption{Five levels of Wavelet Decomposition, with the best metrics marked in \textbf{bold}.}
\centering
\renewcommand{\arraystretch}{1.6} 
\scriptsize
\label{wavelet_level}
\resizebox{1.0\columnwidth}{!}{
\begin{tabular}{c|llllcccc}
\hline
     & \multicolumn{4}{c}{BeltCrack14ks}    & \multicolumn{4}{c}{BeltCrack9kd}  \\ \cline{2-9} 
\multirow{-2}{*}{Level} & \multicolumn{1}{c}{mAP50} & \multicolumn{1}{c}{Precisoin} & \multicolumn{1}{c}{Recall} & \multicolumn{1}{c|}{F1}    & mAP50                         & Precisoin                                                         & Recall                                                            & F1                                                                \\ \hline
1                          & 64.76                     & 90.52                         & 72.78                      & \multicolumn{1}{l|}{80.69} & 19.08                         & 42.14                                                             & 40.33                                                             & 41.56                                                             \\
2                          & 64.93                     & 89.61                         & 73.69                      & \multicolumn{1}{l|}{80.87} & 23.52 & 47.89 &  48.12 &  48.01 \\
3                          & \textbf{67.08}                     & \textbf{91.03}                        & \textbf{74.89}                     & \multicolumn{1}{l|}{\textbf{82.17}} & \textbf{30.79} & \textbf{61.69}         & \textbf{54.94}         & \textbf{56.98}              \\
4                          & 65.26                     & 89.53                         & 73.72                      & \multicolumn{1}{l|}{81.13} & 24.15 &  49.32 & 51.67 & 50.47 \\
5                          & 64.31                     & 88.87                         & 73.62                      & \multicolumn{1}{l|}{80.51} & 20.83 & 45.68 & 46.29 &  45.98 \\ \hline
\end{tabular}}
\end{table}

\begin{figure}
	\centering
	\includegraphics[width=1.0\columnwidth]{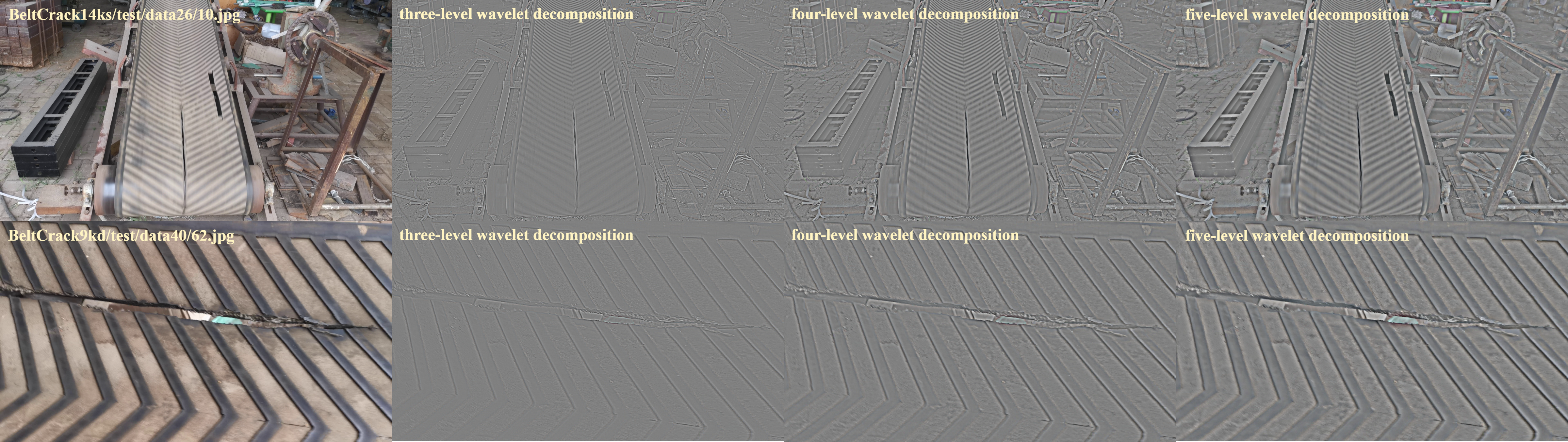}
    \caption{The effects of different-level wavelet decomposition on our detection performance.}
	\label{fig:wavelet_level_fig}
\end{figure}

As shown in Table \ref{wavelet_level} and Fig. \ref{fig:wavelet_level_fig}, performance varies with the decomposition level. When Wavelet level increases from 1 to 3, all metrics improve consistently, mAP$_{50}$ rising from 64.76 to 67.08 and F1 rising from 80.69 to 82.17. This implies that a moderate decomposition level can enhance the model's ability to capture crack textures and suppress background noise by enriching multi-scale representations. However, increasing the level beyond 3 doesn't bring further gains and even leads to a slight decline, implying that higher decomposition levels may cause redundancy and weaken the high-frequency details useful for localization. Therefore, a 3-level decomposition provides an effective balance between feature enrichment and computational efficiency.

\section{Conclusions}
To advance the intelligent development of learning-based belt crack detection, 
this paper constructs  the first pair of sequential datasets specifically for industrial belt crack detection. Furthermore, beyond traditional spatial and spatio-temporal modeling, it proposes a baseline model, $i.e.$, BeltCrackDet, with triple-domain feature learning.
Extensive experiments demonstrate that: (i) this pair of datasets is highly usable and effective for learning-based belt crack detection, and (ii) the proposed baseline is clearly superior to other compared methods, often achieving SOTA detection performance on both datasets. Despite its superior accuracy, the baseline still exhibits relatively high network complexity and low inference speed. These weaknesses could impact the real-time applicability in industrial environments. Therefore, future research should focus on optimizing the network architecture of our BeltCrackDet to reduce computational cost, enabling broader deployment in real-world industrial belt inspection systems.

\bibliographystyle{unsrt}

\bibliography{crack}

\end{document}